\documentclass[sigconf]{acmart}

\usepackage{graphicx}
\usepackage{subcaption}
\usepackage{booktabs}
\usepackage{multirow}
\usepackage{adjustbox}
\usepackage{amsmath}
\usepackage{verbatim} 

\AtBeginDocument{%
  }

\renewcommand\footnotetextcopyrightpermission[1]{}

\begin{document}

\title{Breaking the Rigid Prior:\texorpdfstring{\\}{} Towards Articulated 3D Anomaly Detection}

\author{Jinye Gan}
\authornote{These authors contributed equally.}
\affiliation{%
  \institution{ShanghaiTech University}
  \country{China}
}
\email{ganjy2024@shanghaitech.edu.cn}

\author{Bozhong Zheng}
\authornotemark[1]
\affiliation{%
  \institution{ShanghaiTech University}
  \country{China}
}
\email{zhengbzh2023@shanghaitech.edu.cn}

\author{Xiaohao Xu}
\authornotemark[1]
\affiliation{%
  \institution{University of Michigan, Ann Arbor}
  \country{USA}
}
\email{xiaohaox@umich.edu}

\author{Junye Ren}
\affiliation{%
  \institution{ShanghaiTech University}
  \country{China}
}
\email{renjy2025@shanghaitech.edu.cn}

\author{Zixuan Zhang}
\affiliation{%
  \institution{ShanghaiTech University}
  \country{China}
}
\email{zhangzx2024@shanghaitech.edu.cn}

\author{Na Ni}
\authornote{Correspondence: Na Ni and Yingna Wu.}
\affiliation{%
  \institution{ShanghaiTech University}
  \country{China}
}
\email{nina@shanghaitech.edu.cn}

\author{Yingna Wu}
\authornotemark[2]
\affiliation{%
  \institution{ShanghaiTech University}
  \country{China}
}
\email{wuyn@shanghaitech.edu.cn}

\renewcommand{\shortauthors}{Gan et al.}

\begin{abstract}
\textcolor{black}{Existing 3D anomaly detection methods are built on a rigid prior: normal geometry is pose-invariant and can be canonicalized through registration or alignment. This prior does not hold for articulated objects
with hinge or sliding joints, where valid pose changes induce structured geometric variations that cannot be collapsed to a single canonical template, causing pose-induced deformations to be 
misidentified as anomalies while true structural defects are obscured.
No existing benchmark addresses this challenge. We introduce \textbf{ArtiAD}, the first large-scale benchmark for articulated 3D anomaly detection, comprising 15,229 point clouds across 39 object categories with dense joint-angle variations and six structural anomaly types. Each sample is annotated with its joint configuration and part-level motion labels, enabling explicit disentanglement of pose-induced geometry from structural defects. ArtiAD also provides a seen/unseen articulation split to evaluate both interpolation and extrapolation to novel joint configurations. We propose Shape--Pose-Aware Signed Distance Field (\textbf{SPA-SDF}), a baseline that replaces the rigid prior with a continuous pose-conditioned implicit field, factorized into an articulation-independent structural prior and a Fourier-encoded joint embedding. At inference, the articulation state is recovered by
minimizing reconstruction energy, and anomalies are identified as point-wise deviations from the learned manifold. SPA-SDF achieves 0.884 object-level AUROC on seen configurations and 0.874 on unseen
configurations, substantially outperforming all rigid-based baselines. Our code and benchmark will be publicly released to facilitate future research. }
\end{abstract}



\begin{teaserfigure}
  \centering
  \includegraphics[width=\textwidth]{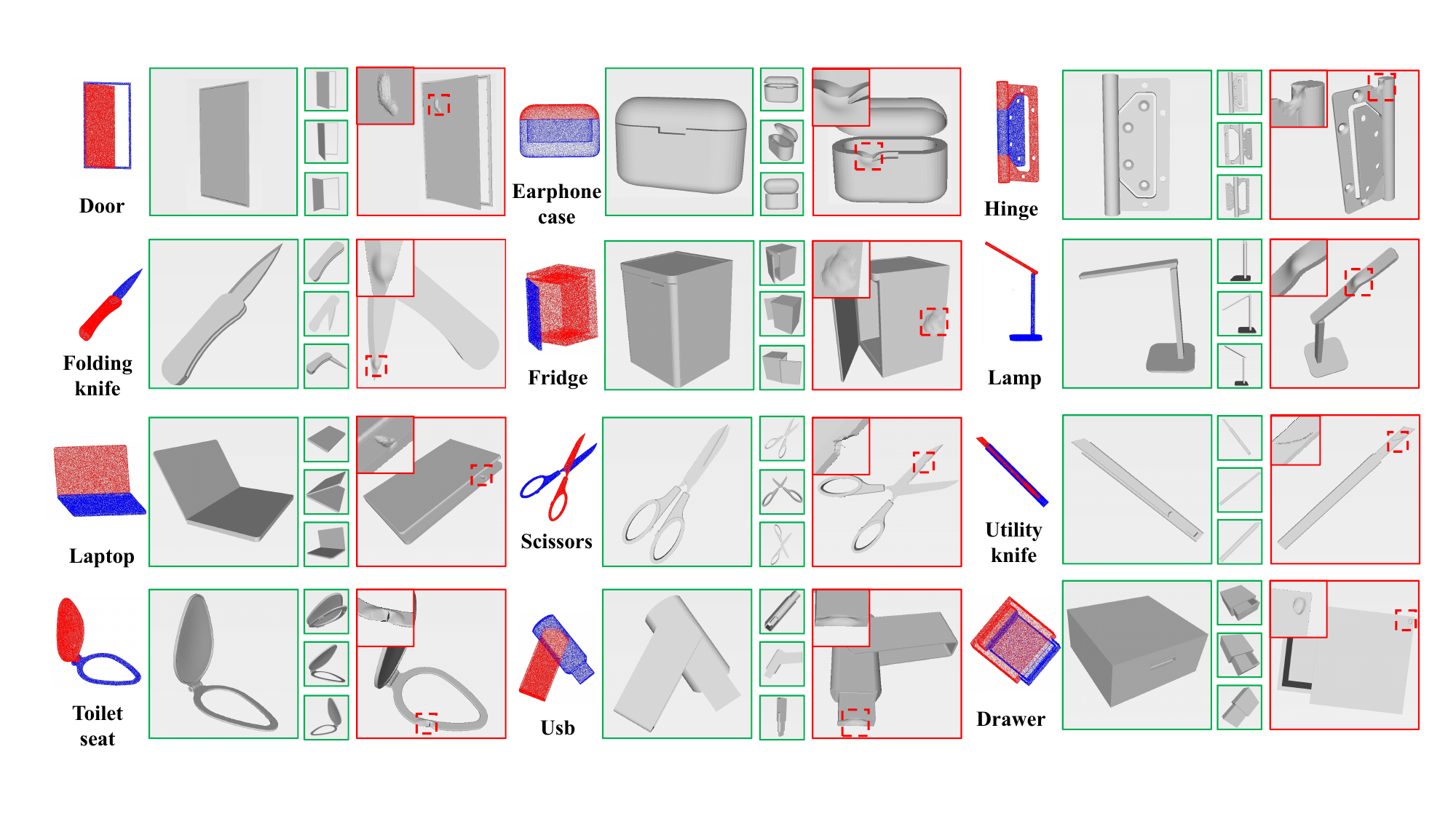}
  \vspace{-13mm}
  \caption{
  Overview of \textbf{ArtiAD}, a large-scale benchmark for articulated 3D anomaly detection.
  The dataset covers diverse articulated object categories with substantial pose variations,
  including hinged and prismatic structures. For each category, we show representative shapes
  together with local structural anomalies on moving parts, highlighting the challenge of
  disentangling articulation-induced geometric variation from genuine defects.
  }
  \Description{Overview of the ArtiAD dataset with multiple articulated object categories and representative local anomalies.}
  \label{fig:overview}\vspace{5mm}
\end{teaserfigure}

\maketitle


\section{Introduction}
\label{sec:intro}

\textcolor{black}{3D anomaly detection is a core component of industrial inspection, where the objective is to identify structural defects that deviate from the normal geometry of manufactured objects. Compared to 2D image-based methods, 3D point clouds encode explicit surface geometry that is invariant to illumination and surface texture, making them particularly effective for detecting fine-grained deformations and structural irregularities~\cite{Bergmann_2022_mvtec_3dad,Real3d-AD}. A broad range of
methods has been proposed to model normal 3D shapes and score geometric deviations through reconstruction~\cite{huang2022registration,R3D-AD,Cheng_2025}, feature matching~\cite{M3DM,Defard_2021_PaDiM,zhou2024pointad,Cao_2024}, or implicit representations~\cite{park2019deepsdf,zheng2025bridging}.}

\textcolor{black}{A foundational assumption shared by virtually all existing 3D anomaly detection methods is what we term the \emph{rigid prior}: normal geometry is pose-invariant, and samples of the same category can be
registered to a single canonical template so that any residual geometric discrepancy reflects a structural defect. This prior is well justified for the objects featured in current benchmarks~\cite{Bergmann_2022_mvtec_3dad,Real3d-AD,IMRNet,zheng2025bridging}, such as bearings, bottle caps, and machined castings, which exhibit negligible pose
variation.}

\textcolor{black}{As illustrated in Fig.~\ref{fig:overview}, the rigid prior breaks for \emph{articulated objects}, whose parts are connected by hinge, revolute, or prismatic joints. Representative examples include doors,
drawers, folding knives, laptops, and injectors. For such objects, different joint configurations produce geometrically distinct yet structurally valid shapes governed by kinematic constraints. Point clouds at different articulation states cannot be registered to any single canonical geometry, because joint motion is an intrinsic degree of freedom rather than a nuisance to be removed. Methods that enforce the rigid prior therefore systematically misclassify
articulation-induced geometry changes as anomalies, while genuine structural defects on moving parts may be suppressed by pose variation. This is a fundamental limitation that improved registration
cannot resolve. Despite substantial progress in 3D and dynamics anomaly
detection~\cite{ye2025po3ad,M3DM,cheng2025towards,horwitz2022empirical,R3D-AD,gu2026multi,li2025multi} and articulated shape modeling~\cite{park2019deepsdf,mu2021asdf,Lei2022CaDeX,Wei2024M2V,pumarola2020d,mildenhall2020nerf}, the intersection of these two problems has not been systematically studied. The absence of datasets with dense articulation coverage, explicit pose annotations, and point-level anomaly labels has been the primary obstacle to progress.}

\textcolor{black}{In this paper, we break the rigid prior and take a step toward articulated 3D anomaly detection. We present \textbf{ArtiAD}, the first large-scale benchmark dedicated to this setting. Built on a fully controllable simulation pipeline, ArtiAD spans 39 articulated object categories and 15,229 point clouds, with dense sampling across continuous joint-angle ranges and six structural anomaly types per category. Each point cloud is annotated with its joint configuration and part-level motion labels, providing supervision to explicitly separate articulation-induced geometry from structural defects. ArtiAD introduces a seen/unseen articulation split that enables independent evaluation of pose interpolation and extrapolation, a capability absent from all prior 3D anomaly detection benchmarks.}

\textcolor{black}{To establish a strong baseline, we propose \textbf{SPA-SDF}, a Shape--Pose Aware Signed Distance Field that replaces the rigid prior with a continuous pose-conditioned implicit manifold. SPA-SDF factorizes structural shape priors from joint conditioning, learning an articulation-aware normality model from normal training samples only. At inference, the unknown articulation state is estimated by minimizing reconstruction energy, and anomaly scores are derived as pointwise deviations from this manifold.}

\noindent \textcolor{black}{Our contributions are as follows:}
\begin{itemize}
  \item \textcolor{black}{We identify the rigid prior as a fundamental obstacle to
        anomaly detection on articulated objects, and formally
        characterize why existing methods fail in this setting.}
  \item \textcolor{black}{We present ArtiAD, the first large-scale benchmark for
        articulated 3D anomaly detection, with dense pose variations,
        six anomaly types per category, and a principled seen/unseen
        articulation split.}
  \item \textcolor{black}{We propose SPA-SDF, a pose-conditioned implicit baseline that replaces the rigid prior, and conduct comprehensive experiments
        demonstrating its advantages over rigid-based methods across
        both interpolation and extrapolation regimes.}
\end{itemize}

\section{Related Work}
\label{sec:related}

\subsection{3D Anomaly Detection Datasets}
\label{sec:related_datasets}

\textcolor{black}{Table~\ref{tab:dataset_comparison} compares ArtiAD against prior 3D anomaly detection benchmarks. Existing benchmarks have evolved along two tracks, yet both share a foundational assumption: all normal samples of the same category can be canonicalized to a single rigid template. On the RGB-D side, MVTec 3D-AD~\cite{Bergmann_2022_mvtec_3dad} established the first industrial 3D benchmark with
10 object categories and up to 5 anomaly types captured via depth sensors, while
Eyecandies~\cite{bonfiglioli2022eyecandies} extended the scope to synthetic confectionery under multimodal supervision; both treat object pose as fixed.
On the pure point cloud side, Real3D-AD~\cite{Real3d-AD} raised geometric fidelity with
structured-light scans across 12 categories, and Anomaly-ShapeNet~\cite{IMRNet} pushed scale to 40 synthetic categories with 6 anomaly types per class, though all samples remain pose-normalized. More recent efforts have probed adjacent challenges without relaxing the rigid assumption: MiniShift-3D~\cite{cheng2025towards} introduces mild viewpoint perturbation across 12 rigid categories, and MulSen-AD~\cite{li2025multi} broadens the sensing modalities to RGB, infrared, and
point cloud for 15 rigid objects.
The common thread running through all of these benchmarks is the absence of continuous articulation variation, joint-angle annotations, or part-level motion labels, making it impossible to train or evaluate methods that must disentangle legitimate kinematic deformation from structural defects.
ArtiAD is the first benchmark in which normal geometry is explicitly pose-dependent, directly and systematically challenging the rigid prior that underlies every prior evaluation protocol.}

\subsection{3D Anomaly Detection Methods}
\label{sec:related_methods}

Anomaly detection has been extensively studied in 2D vision,
where methods such as reconstruction-based models and feature matching
approaches have achieved strong performance~\cite{roth2022towards}.
Recent advances extend these ideas to 3D data, leveraging geometric
representations such as point clouds, meshes, and multi-view images~\cite{M3DM,Real3d-AD,chen2026unsupervised}.

Existing 3D anomaly detection methods can be broadly categorized into
feature-based and reconstruction-based approaches.
Feature-based methods extract local geometric descriptors or learned
representations and detect anomalies via distance or memory bank mechanisms.
Representative works include PatchCore-style approaches~\cite{roth2022towards}
adapted to 3D data, as well as M3DM~\cite{M3DM},
which leverages multi-scale point cloud features for anomaly localization.
Other works, such as BTF~\cite{horwitz2022empirical},
utilize handcrafted geometric features (e.g., FPFH) for anomaly detection.
Reconstruction-based methods instead model the distribution of normal data
and identify anomalies via reconstruction errors.
Recent approaches explore implicit representations or generative models
for 3D anomaly detection, aiming to capture the underlying geometric structure~\cite{zheng2025bridging}.
Datasets such as Real3D-AD~\cite{Real3d-AD}
have facilitated benchmarking of these methods on real-world industrial data.

Despite these advances, existing methods generally rely on a rigid prior,
assuming that normal geometry is pose-invariant and can be aligned into
a canonical space.
This assumption breaks down for articulated objects,
where pose variations induce structured geometric changes,
leading to false positives and degraded detection performance.

\begin{table}[t]
    \centering
    \caption{\textcolor{black}{Comparison of 3D anomaly detection datasets. `Syn', `IR', `D', and `PC' denote Synthetic, Infrared, Depth, and Point Cloud, respectively.}} \vspace{-2mm}
    \label{tab:dataset_comparison}
    \footnotesize
    \setlength{\tabcolsep}{3pt}
    \begin{tabular}{lcccccccc}
    \toprule
    Dataset & Year & Type & Modality & Cls & Anom. & Num & Pose \\
    \midrule
    MVTec3D-AD~\cite{Bergmann_2022_mvtec_3dad}     
    & 2021 & Real & RGB-D & 10 & 3--5 & 3604 & $\times$ \\
    
    Eyecandies~\cite{bonfiglioli2022eyecandies}            
    & 2022 & Syn  & RGB-D & 10 & 3 & 15500 & $\times$ \\
    
    Real3D-AD~\cite{Real3d-AD}               
    & 2023 & Real & PC & 12 & 2 & 1200 & $\times$ \\
    
    Anomaly-ShapeNet~\cite{IMRNet} 
    & 2024 & Syn  & PC & 40 & 6 & 1600 & $\times$ \\

    MiniShift~\cite{cheng2025towards}
    & 2026 & Real & PC & 12 & 3 & 2577 & $\times$ \\
    
    MulSen-AD~\cite{li2025multi}                              
    & 2025 & Real & RGB+IR+PC & 15 & 14 & 2035 & $\times$ \\
    
    \midrule
    \textbf{ArtiAD (Ours)}           
    & 2026 & Syn & PC & 39 & 6 & 15229 & $\checkmark$ \\
    \bottomrule
    \end{tabular}\vspace{5mm}
\end{table}

\subsection{Articulated Object Modeling}
\label{sec:related_articulated}

Articulated object modeling has been widely studied for part
segmentation, pose estimation, and shape reconstruction.
SAPIEN~\cite{xiang2020sapien}
provides large-scale datasets of articulated objects with annotated
joint axes and kinematic ranges. ANCSH~\cite{li2020category} estimates
category-level articulation from single observations.
Recent works further explore continuous implicit representations
for articulated geometry, such as PARIS~\cite{liu2023paris},
NAP~\cite{lei2023nap}, and A-SDF~\cite{mu2021asdf},
which model shape variations across articulation states in a unified latent space.
Dynamic neural representations, including D-NeRF~\cite{pumarola2020d}
and Motion2VecSets~\cite{Wei2024M2V}, further enable modeling of
continuous deformation and temporal geometry.
The key distinction from our setting is the downstream objective.
Existing articulated modeling methods treat all observed geometry as
valid, whereas anomaly detection requires identifying localized
deviations from a pose-conditioned normal distribution.
This demands a model that encodes the rigid prior's complement:
a continuous, pose-dependent normality manifold against which
geometric outliers can be scored.

\section{The ArtiAD Benchmark}
\label{sec:dataset}

\begin{figure}[t]
    \centering
    \begin{subfigure}{\linewidth}
        \centering
        \includegraphics[width=\linewidth, trim=40 60 40 80, clip]{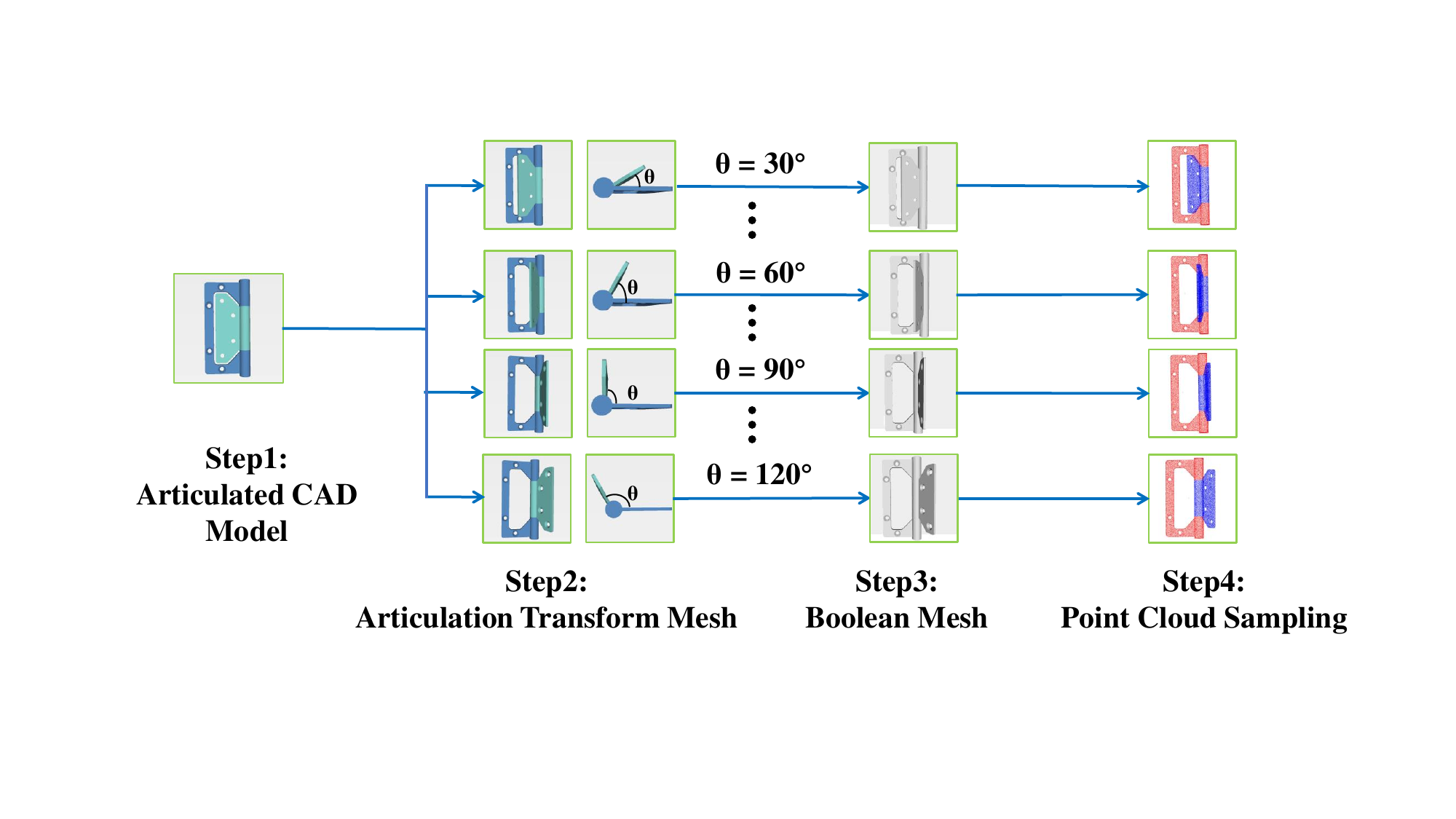}
        \vspace{-10mm}
        \caption{Normal sample generation pipeline}
        \label{fig:pipeline_normal}
    \end{subfigure}
    
    \vspace{10pt}
    
    \begin{subfigure}{\linewidth}
        \centering
        \includegraphics[width=\linewidth, trim=40 60 40 80, clip]{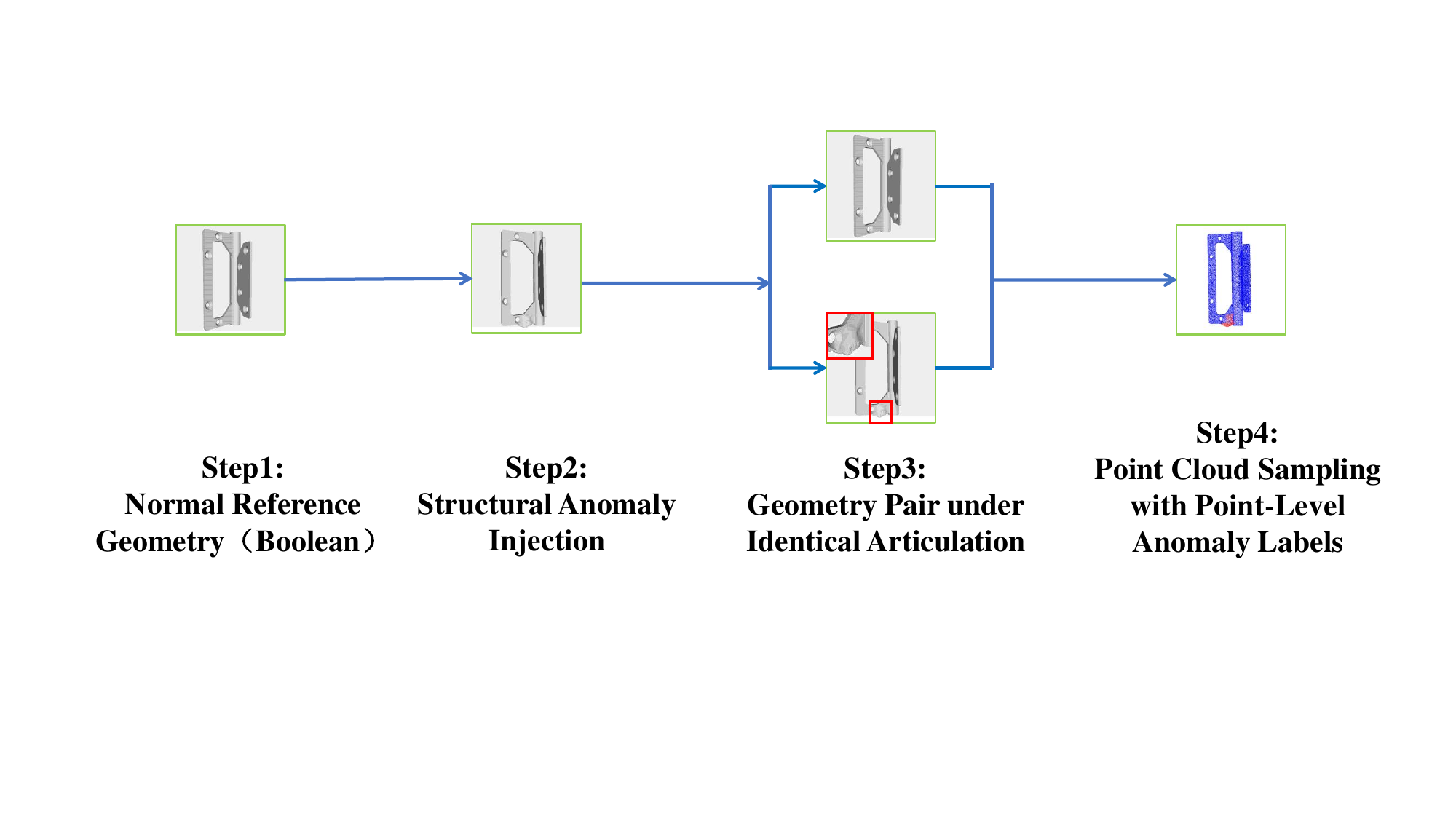}
        \vspace{-15mm}
        \caption{Abnormal sample generation pipeline}
        \label{fig:pipeline_abnormal}
    \end{subfigure}
    
    \caption{Data generation pipelines of ArtiAD. (a) Normal samples are generated from pose-conditioned reference meshes without defect injection. (b) Abnormal samples are created by injecting structural anomalies under identical articulation states, enabling pose-aligned normal--defective pairing and point-level anomaly supervision. }
    \Description{
    Two data generation pipelines of the ArtiAD dataset. 
    The first shows normal samples generated from articulated meshes without defects, 
    and the second shows abnormal samples with injected structural anomalies under the same articulation states.
    }
    \label{fig:pipeline_total}\vspace{5mm}
\end{figure}

\subsection{Data Generation Pipeline}
\label{sec:pipeline}

\begin{figure*}[t]
    \centering
    \includegraphics[width=\textwidth]{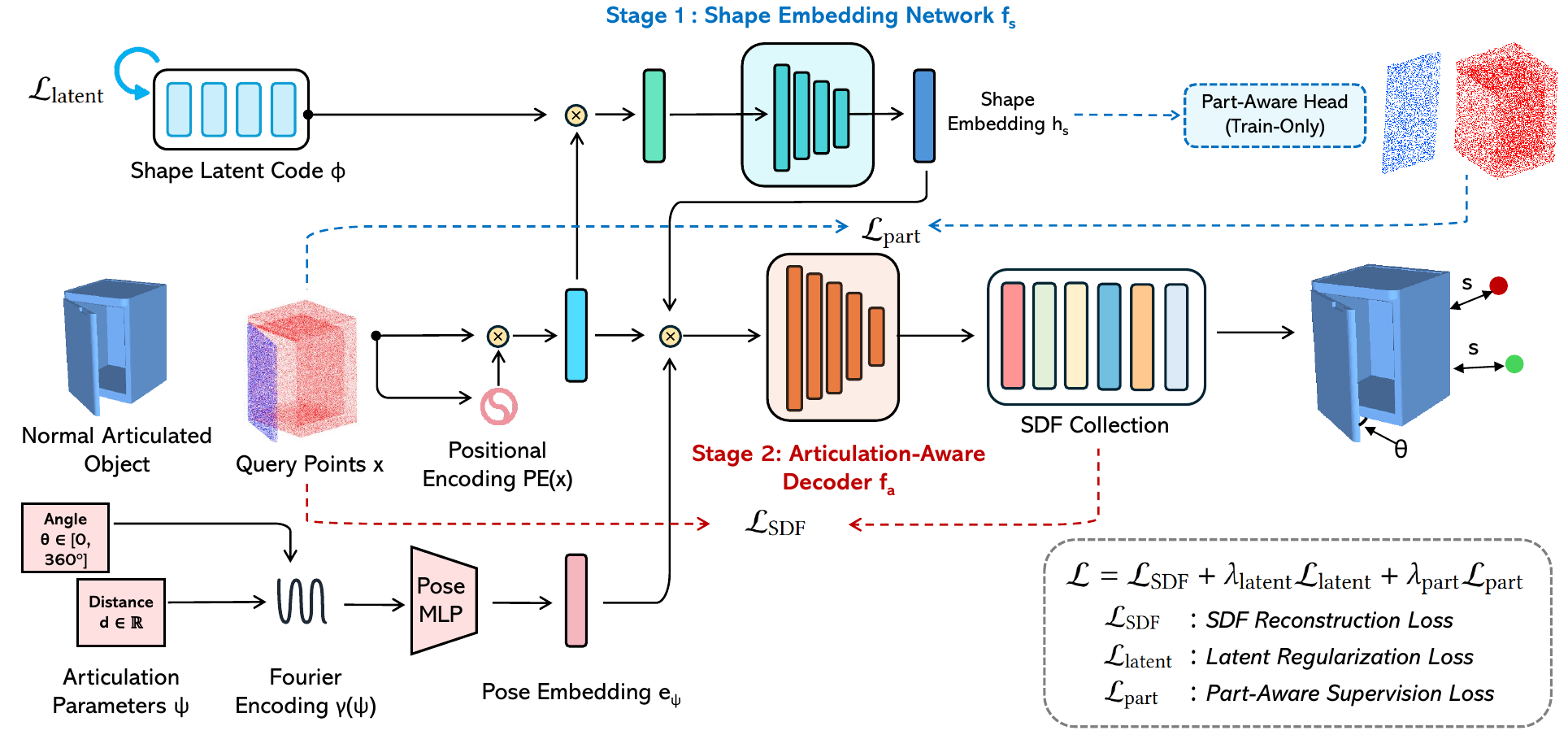}
    \caption{Overview of Shape--Pose-Aware Signed Distance Field (\textbf{SPA-SDF}) framework.
    Given an input point cloud of an articulated object, our method models normal geometry using a pose-conditioned signed distance field $f(\mathbf{x}; \phi, \psi)$.
    The framework factorizes structural shape and articulation by combining a category-level latent code $\phi$ with Fourier-encoded pose features $\gamma(\psi)$.
    In the network, spatial query points are first processed by a shape embedding network to obtain articulation-independent structural features, which are then fused with pose encoding in an articulation-aware decoder to predict signed distances.
    During training, the model is optimized with SDF reconstruction, latent regularization, and part-aware supervision.
    At inference, the articulation parameter is estimated by minimizing reconstruction energy, and anomaly scores are computed from the residual field, enabling robust anomaly localization under diverse pose variations.
    }
    \label{fig:pipeline}\vspace{5mm}
\end{figure*}

\textcolor{black}{ArtiAD is constructed through a fully controllable synthetic pipeline
that provides explicit control over joint kinematics, part-level
geometry, and defect injection. This simulation-based approach
guarantees disentangled point-level supervision over both articulation
state and anomaly type. As shown in Fig.~\ref{fig:pipeline}, two
complementary sub-pipelines generate normal and abnormal samples under
identical articulation states.}

\paragraph{Articulated Object Modeling.}
\textcolor{black}{We begin by selecting articulated objects with clear kinematic structures, including hinge-based and sliding-joint mechanisms commonly found in industrial and daily-use objects. Each object is decomposed into motion-consistent parts according to its physical articulation, and joint axes or sliding directions are manually specified. For hinge-type objects, a rotational joint parameter $\theta$ defines the relative pose between moving and static parts, while sliding mechanisms are governed by a translation parameter $d$. These parameters are treated as continuous variables, forming a low-dimensional articulation space:
\begin{equation}
\mathbf{p} = \{\theta \ \text{or} \ d\},
\end{equation}
where each configuration corresponds to a valid geometry.}

\paragraph{Pose-Conditioned Normal Reference Geometry.}
\textcolor{black}{For each object category, articulation parameters are densely sampled from the continuous kinematic space to cover the full motion range. Given articulation parameter $\mathbf{p}$, part transformations are applied through rigid-body motion to obtain a pose-conditioned reference mesh:
\begin{equation}
\mathcal{M}(\mathbf{p}) = \mathcal{T}(\mathbf{p}) \big( \mathcal{M}_{\text{parts}} \big).
\end{equation}
These meshes represent geometrically distinct yet structurally valid shapes. Because articulation alters global geometry in a structured manner, these variations cannot be removed by canonical alignment, making them fundamentally different from rigid pose changes.}

\paragraph{Structural Anomaly Injection.}
\textcolor{black}{Abnormal samples are generated from the same pose-conditioned reference meshes but undergo additional structural anomaly injection. Localized geometric perturbations simulate six defect types: dents, bulges, fractures, bending deformations, surface distortions, and missing material. The injection of anomaly \emph{occurs after} the articulation transformation, ensuring that the pose-induced and defect-induced geometry coexist but originate from independent generative factors.}

\paragraph{Pose-Aligned Pairing.}
\textcolor{black}{For each articulation state $\mathbf{p}$, paired normal and defective
meshes are constructed under identical poses. Geometric discrepancies
between paired samples therefore arise exclusively from injected
structural defects, enabling precise point-level supervision.}

\paragraph{Point Cloud Sampling and Annotation.}
\textcolor{black}{All meshes are converted to point clouds by uniform area-weighted
surface sampling at 16384 points per cloud, normalized to a consistent
scale and coordinate frame. Part-level motion labels are propagated via
nearest-neighbor lookup. For abnormal samples, points in defect regions
receive binary anomaly masks. Ground-truth signed distances are stored
to support SDF-based training.}

\begin{table}[t]
\centering
\caption{Dataset statistics of SPA-SDF. All categories contain 6 anomaly types.}\vspace{-2mm}
\label{tab:dataset_statistics}
\footnotesize
\setlength{\tabcolsep}{1.8pt}

\begin{tabular}{lcccc|lcccc}
\toprule
\textbf{Category} & \textbf{Train} & \textbf{Seen} & \textbf{Unseen} & \textbf{Total}
& \textbf{Category} & \textbf{Train} & \textbf{Seen} & \textbf{Unseen} & \textbf{Total} \\
\midrule

Hinge & 320 & 83 & 90 & 493
& Door & 151 & 71 & 69 & 291 \\
Hinge2 & 320 & 84 & 84 & 488
& Door2 & 271 & 70 & 70 & 411 \\
Hinge3 & 320 & 84 & 84 & 488
& Door3 & 271 & 70 & 70 & 411 \\

Drawer & 141 & 97 & 96 & 334
& Earphone & 241 & 70 & 70 & 381 \\
Drawer2 & 141 & 70 & 70 & 281
& Earphone2 & 241 & 70 & 70 & 381 \\
Drawer3 & 221 & 70 & 70 & 361
& Earphone3 & 241 & 70 & 70 & 381 \\

Knife & 320 & 82 & 84 & 486
& Fridge & 150 & 71 & 72 & 293 \\
Knife2 & 271 & 70 & 70 & 411
& Fridge2 & 210 & 72 & 74 & 356 \\
Knife3 & 331 & 70 & 70 & 471
& Fridge3 & 271 & 70 & 70 & 411 \\

Injector & 113 & 70 & 69 & 252
& Lamp & 331 & 70 & 70 & 471 \\
Injector2 & 316 & 70 & 70 & 456
& Lamp2 & 331 & 70 & 70 & 471 \\
Injector3 & 316 & 70 & 69 & 455
& Lamp3 & 331 & 70 & 70 & 471 \\

Laptop & 210 & 70 & 72 & 352
& Scissors & 191 & 70 & 70 & 331 \\
Laptop2 & 210 & 70 & 70 & 350
& Scissors2 & 191 & 70 & 70 & 331 \\
Laptop3 & 210 & 70 & 70 & 350
& Scissors3 & 191 & 70 & 70 & 331 \\

Seat & 167 & 74 & 78 & 319
& USB & 320 & 92 & 92 & 504 \\
Seat2 & 211 & 70 & 70 & 351
& USB2 & 320 & 70 & 70 & 460 \\
Seat3 & 211 & 70 & 70 & 351
& USB3 & 320 & 70 & 70 & 460 \\

UtilKnife & 150 & 70 & 72 & 292
& 
UtilKnife2 & 171 & 70 & 70 & 311
\\
UtilKnife3 & 291 & 70 & 70 & 431
& -- & -- & -- & -- & -- \\

\midrule
\textbf{Mean} & 244.46 & 72.82 & 73.21 & 390.49
& \textbf{Total} & 9534 & 2840 & 2855 & 15229 \\
\bottomrule
\end{tabular}\vspace{5mm}
\end{table}

\subsection{Dataset Statistics and Splits}
\label{sec:stats}

\textcolor{black}{ArtiAD contains 15,229 point clouds across 39 articulated object
categories, as detailed in Table~\ref{tab:dataset_statistics}. The
dataset covers two fundamental mechanisms: hinge-based rotation
(doors, hinges, laptops, scissors, folding knives) and prismatic
sliding (drawers, injectors, USB connectors).}

\paragraph{Articulation-Space Splits.}\textcolor{black}{
ArtiAD is organized by articulation state rather than by instance. The
\textbf{Train} split provides normal samples at a subset of joint
configurations, used to learn the pose-dependent normality manifold.
The \textbf{Seen} split evaluates anomaly detection within the
training articulation range, measuring interpolation accuracy. The
\textbf{Unseen} split evaluates configurations outside the training
range, measuring extrapolation to novel joint states. This two-level
protocol directly quantifies whether a method has broken the rigid
prior or merely approximated it within a narrow pose range.}

\paragraph{Anomaly Coverage.}
\textcolor{black}{All 39 categories include the same six anomaly types, injected after
articulation transformation to ensure anomaly labels are independent
of pose. For 13 base object classes, three geometric variants are
provided (e.g., Hinge, Hinge2, Hinge3), introducing intra-class shape
diversity while preserving kinematic structure.}

\section{Shape--Pose-Aware Signed Distance Field (SPA-SDF)}
\label{sec:method}

\subsection{Problem Formulation}
\label{sec:problem}

\textcolor{black}{Let $\mathbf{X} \in \mathbb{R}^{N \times 3}$ denote a point cloud
sampled from an articulated object at joint configuration $\psi \in
\Psi$, where $\Psi$ is the continuous articulation space. Under the
rigid prior, all normal samples are assumed to share a single
canonical geometry. We replace this with a \emph{pose-dependent shape
manifold}:
\begin{equation}
  \mathcal{M} = \{\,\mathbf{X}(\psi) \mid \psi \in \Psi\,\},
  \label{eq:manifold}
\end{equation}
where each $\psi$ corresponds to a physically valid geometry.
Structural anomalies introduce localized perturbations $\Delta$
independent of articulation:
\begin{equation}
  \mathbf{X}_{\text{abn}} = \mathbf{X}(\psi) + \Delta.
  \label{eq:anomaly}
\end{equation}
During training, only normal samples are available, with articulation
states drawn from $\Psi_{\text{train}} \subset \Psi$. At test time,
$\psi$ is unknown. Evaluation is conducted under two regimes: the
\emph{seen} regime ($\psi \in \Psi_{\text{train}}$, interpolation)
and the \emph{unseen} regime
($\psi \in \Psi \setminus \Psi_{\text{train}}$, extrapolation). The
goal is to determine whether a test point cloud lies on $\mathcal{M}$,
robustly across both regimes.}

\subsection{\textcolor{black}{Shape--Pose Factorized Representation}}
\label{sec:representation}

\textcolor{black}{SPA-SDF replaces the rigid prior with a pose-conditioned signed
distance field:
\begin{equation}
  \hat{s} = f(\mathbf{x};\,\phi,\,\psi),
  \label{eq:sdf}
\end{equation}
where $\mathbf{x} \in \mathbb{R}^3$ is a spatial query point, $\phi
\in \mathbb{R}^d$ is a category-level shape latent code capturing
articulation-independent structural priors, and $\psi$ encodes the
joint configuration. This factorization enables disentangled modeling
of structural shape and pose-dependent geometry.}

\paragraph{Articulation Encoding.}
\textcolor{black}{To represent periodic and multi-scale motion patterns, $\psi$ is
encoded with multi-frequency Fourier features:
\begin{equation}
  \gamma(\psi) =
  \bigl[\sin(2^k\psi),\,\cos(2^k\psi)\bigr]_{k=0}^{L-1}.
  \label{eq:fourier}
\end{equation}
This encoding captures fine-grained geometry differences across
joint-angle frequencies without overfitting to the mean shape. The
ablation in Table~\ref{tab:pose_ablation} confirms that Fourier encoding
consistently outperforms using $\psi$ directly.}

\paragraph{SDF Prediction.}
\textcolor{black}{The signed distance is predicted by conditioning jointly on structural
and articulation factors:
\begin{equation}
  \hat{s} = f(\mathbf{x};\,\phi,\,\gamma(\psi)).
  \label{eq:sdf_conditioned}
\end{equation}
This enables SPA-SDF to represent a continuous, pose-dependent
normality manifold in place of the single rigid template assumed by
prior methods.}

\subsection{Network Architecture}
\label{sec:architecture}

\textcolor{black}{SPA-SDF is a two-stage implicit network. Spatial query points are
first encoded with multi-frequency positional encoding:
\begin{equation}
  \mathrm{PE}(\mathbf{x}) =
  \bigl[\sin(2^k\pi\mathbf{x}),\,\cos(2^k\pi\mathbf{x})\bigr]_{k=0}^{K-1}.
\end{equation}}

\paragraph{Stage 1: Shape Embedding Network $f_s$.}
\textcolor{black}{This stage maps spatial coordinates and the shape latent code to a
structure-aware embedding:
\begin{equation}
  \mathbf{h}_s =
  f_s\!\left(\mathbf{x},\,\mathrm{PE}(\mathbf{x}),\,\phi\right).
  \label{eq:shape_embed}
\end{equation}
Skip connections preserve low-level geometric detail. The output
$\mathbf{h}_s$ encodes structural information shared across all
articulation states, corresponding to the articulation-independent
component of the factorized representation.}

\paragraph{Stage 2: Articulation-Aware Decoder $f_a$.}
\textcolor{black}{The decoder predicts the signed distance from both structural and
articulation features:
\begin{equation}
  \hat{s} =
  f_a\!\left(\mathbf{x},\,\mathrm{PE}(\mathbf{x}),\,
  \mathbf{h}_s,\,\mathbf{e}_\psi\right),
  \label{eq:decoder}
\end{equation}
where $\mathbf{e}_\psi$ is obtained by applying a two-layer MLP to
$\gamma(\psi)$. This two-stage design explicitly separates structural
encoding from articulation conditioning, enabling the field to vary
smoothly and physically across the joint space.}

\paragraph{Part-Aware Auxiliary Head.}\textcolor{black}{
An auxiliary part classification head on $\mathbf{h}_s$ is used
during training as a regularization signal to encourage geometrically
consistent embeddings across articulation states. It is discarded at
inference.}

\subsection{Training Objective}

\textcolor{black}{SPA-SDF is trained using only normal samples.
Given a training point cloud $\mathbf{X}(\psi)$ with articulation parameter $\psi$,
we supervise the signed distance prediction at sampled spatial points.}

\paragraph{SDF Reconstruction Loss.}
\textcolor{black}{For each query point $\mathbf{x}_i$ with ground-truth signed distance $s_i$, the primary objective minimizes the L1 reconstruction error:
\begin{equation}
\mathcal{L}_{\text{SDF}} 
= \frac{1}{N} \sum_{i=1}^{N} 
\left| f(\mathbf{x}_i; \phi, \psi) - s_i \right|.
\end{equation}}

\paragraph{Latent Regularization.}\textcolor{black}{
To prevent the category-level latent code from drifting to arbitrarily large values,
we apply an $\ell_2$ regularization term:
\begin{equation}
\mathcal{L}_{\text{latent}} 
= \|\phi\|_2^2.
\end{equation}}

\paragraph{Part-Aware Regularization.}
\textcolor{black}{To encourage geometrically consistent embeddings across articulation states,
we introduce an auxiliary part classification objective on the intermediate shape embedding.
Let $p_i$ denote the ground-truth part label of point $\mathbf{x}_i$.
We apply a cross-entropy loss:
\begin{equation}
\mathcal{L}_{\text{part}} 
= - \sum_{i} p_i \log \hat{p}_i,
\end{equation}
where $\hat{p}_i$ is the predicted part probability.}

\paragraph{Overall Objective.}
\textcolor{black}{The final training objective is:
\begin{equation}
\mathcal{L} 
= \mathcal{L}_{\text{SDF}} 
+ \lambda_{\text{latent}} \mathcal{L}_{\text{latent}}
+ \lambda_{\text{part}} \mathcal{L}_{\text{part}},
\end{equation}
where $\lambda_{\text{latent}}$ and $\lambda_{\text{part}}$ are balance coefficients.}

\subsection{\textcolor{black}{Inference: Articulation Estimation and Anomaly Scoring}}
\label{sec:inference}

\paragraph{Articulation Estimation.}
\textcolor{black}{At inference, $\psi$ is unknown. We recover it by minimizing the implicit reconstruction energy:
\begin{equation}
  \psi^* = \arg\min_{\psi \in \Psi}\,
  \frac{1}{N}\sum_{i=1}^{N}
  \bigl|f(\mathbf{x}_i;\,\phi,\,\psi)\bigr|.
  \label{eq:pose_est}
\end{equation}
For a normal sample, the minimizing $\psi^*$ aligns the observation
with the learned manifold, yielding low reconstruction energy. For an
anomalous sample, structural deviations cannot be explained by
articulation alone, resulting in persistently elevated energy across
all $\psi$. The articulation space is one-dimensional for all ArtiAD
categories, making exhaustive grid search computationally efficient
and immune to local minima.}

\paragraph{Anomaly Scoring.}
\textcolor{black}{Given $\psi^*$, the pointwise anomaly score is:
\begin{equation}
  A(\mathbf{x}_i) =
  \bigl|f(\mathbf{x}_i;\,\phi,\,\psi^*)\bigr|.
  \label{eq:score}
\end{equation}
The object-level score is the mean of the top-$k$ pointwise scores,
providing robustness to surface-sampling noise while retaining
sensitivity to localized defects.}

\section{Experiments}

\subsection{Experimental Setup}
\label{sec:setup}

\paragraph{Baselines.}
\textcolor{black}{We evaluate eight baselines spanning the major paradigms under the
rigid prior: BTF (Raw) and BTF (FPFH)~\cite{horwitz2022empirical}, M3DM (PointBERT)
and M3DM (Point-MAE)~\cite{M3DM}, PatchCore (FPFH) and PatchCore
(Point-MAE)~\cite{roth2022towards}, PO3AD~\cite{ye2025po3ad}, Reg3D-AD~\cite{Real3d-AD} and
PASDF~\cite{zheng2025bridging}. All baselines are trained from scratch on the
ArtiAD training split under identical data conditions.}

\paragraph{Implementation.}
\textcolor{black}{SPA-SDF uses a shape latent code of dimension $d{=}256$, $L{=}16$ Fourier frequencies for articulation encoding, and $K{=}10$ frequencies for spatial positional encoding. We set $\lambda_{\text{latent}}{=}10^{-4}$
and $\lambda_{\text{part}}{=}0.1$. Each category is trained
independently for 800 epochs using Adam with learning rate
$10^{-4}$.}

\subsection{Evaluation Metrics}
\label{sec:metrics}

\textcolor{black}{Following existing 3D anomaly detection benchmarks~\cite{Real3d-AD,Bergmann_2022_mvtec_3dad}, we report two metrics at two levels of granularity. \textbf{Point-level AUROC} measures localization accuracy by evaluating point-wise anomaly scores against binary ground-truth
masks. \textbf{Object-level AUROC} measures classification accuracy by aggregating point-wise scores into a single object-level score (mean of top-$k$ scores) and computing the AUROC over normal/abnormal labels. Both metrics are averaged over the 39 categories and reported separately for seen and unseen splits.}

\subsection{Benchmarking Results on ArtiAD}
\label{sec:main_results}

Table~\ref{tab:overall_comparison_summary} reports mean Pt-AUROC and Obj-AUROC across all
39 categories on both splits. Per-category results appear in
Tables~\ref{tab:benchmark_seen}.

\paragraph{Methods enforcing the rigid prior fail systematically.}\textcolor{black}{
BTF achieves near-random Obj-AUROC (0.54 to 0.59 across both splits),
consistent with the expectation that a fixed-pose memory bank cannot
separate kinematic deformation from structural defects. PatchCore
performs similarly (0.56 to 0.64). PO3AD, which incorporates
part-aware registration, reaches only 0.45 to 0.47 Obj-AUROC,
indicating that forcing articulated samples into a rigid canonical
frame actively degrades performance. \textcolor{black}{
Reg3D-AD achieves moderate performance (0.74 to 0.76), suggesting that while explicit 
geometric modeling and registration improve over purely memory-based 
methods, the reliance on rigid alignment still limits robustness under 
large articulation variations.} PASDF achieves 0.68 to 0.69,
suggesting that implicit representations offer some robustness, but
the rigid prior embedded in its training still limits detection
accuracy on pose-varying categories.}

\paragraph{Feature-learning methods are partially resilient but limited.}
\textcolor{black}{M3DM (Point-MAE) achieves the strongest baseline performance (0.878
and 0.871 Obj-AUROC on seen and unseen splits), demonstrating the
value of large-scale pretrained point cloud features. However, the
method implicitly assumes fixed feature distributions and shows a
non-trivial false positive rate on categories with large kinematic
ranges, such as folding knives and scissors
(as shown in Table~\ref{tab:benchmark_seen}).}

\begin{table}[t]
    \centering
    \caption{Overall comparison on the ArtiAD dataset. We report mean point-level (Pt) and object-level (Obj) AUROC over all 39 categories for seen and unseen splits.} \vspace{-2mm}
    \label{tab:overall_comparison_summary}
    \footnotesize
    \setlength{\tabcolsep}{5pt}
    \begin{tabular}{lcccc}
    \toprule
    \textbf{Method} 
    & \textbf{Seen-Pt} 
    & \textbf{Seen-Obj} 
    & \textbf{Unseen-Pt} 
    & \textbf{Unseen-Obj} \\
    \midrule
    BTF (Raw)             & 0.5379 & 0.5416 & 0.5485 & 0.5522 \\
    BTF (FPFH)            & 0.5770 & 0.5893 & 0.5746 & 0.5880 \\
    M3DM (Point-BERT)     & 0.7046 & 0.8572 & 0.7074 & 0.8655 \\
    M3DM (Point-MAE)      & 0.7452 & 0.8782 & 0.7468 & 0.8714 \\
    PatchCore (FPFH)      & 0.5019 & 0.6397 & 0.5090 & 0.6424 \\
    PatchCore (Point-MAE) & 0.5043 & 0.5892 & 0.4828 & 0.5636 \\
    Reg3D-AD                 & 0.6502 & 0.7479 & 0.6531 & 0.7663 \\
    PO3AD                 & 0.5035 & 0.4453 & 0.4979 & 0.4672 \\
    PASDF                 & 0.6620 & 0.6840 & 0.6950 & 0.6890 \\
    \midrule
    \textbf{SPA-SDF (Ours)} & \textbf{0.8651} & \textbf{0.8843} & \textbf{0.8410} & \textbf{0.8740} \\
    \bottomrule
    \end{tabular}\vspace{7mm}
\end{table}
\begin{table*}[t]
\centering
\caption{Comparison on seen classes. We report point-level AUROC (Pt) and object-level AUROC (Obj).} \vspace{-2mm}
\label{tab:benchmark_seen}
\footnotesize
\setlength{\tabcolsep}{2.2pt}
\begin{adjustbox}{width=\textwidth}
\begin{tabular}{l*{10}{cc}}
\toprule
\multirow{2}{*}{Method}
& \multicolumn{2}{c}{hinge}
& \multicolumn{2}{c}{door}
& \multicolumn{2}{c}{drawer}
& \multicolumn{2}{c}{earphone\_case}
& \multicolumn{2}{c}{fridge}
& \multicolumn{2}{c}{injector}
& \multicolumn{2}{c}{lamp}
& \multicolumn{2}{c}{laptop}
& \multicolumn{2}{c}{scissors}
& \multicolumn{2}{c}{toilet\_seat} \\
\cmidrule(lr){2-21}
& Pt & Obj & Pt & Obj & Pt & Obj & Pt & Obj & Pt & Obj
& Pt & Obj & Pt & Obj & Pt & Obj & Pt & Obj & Pt & Obj \\
\midrule
BTF (Raw)
& 0.4854 & 0.4949
& 0.5820 & 0.4525
& 0.5079 & 0.9154
& 0.5420 & 0.6833
& 0.4560 & 0.4623
& 0.5655 & 0.4071
& 0.5390 & 0.6433
& 0.5384 & 0.5250
& 0.4972 & 0.6350
& 0.5113 & 0.4719 \\

BTF (FPFH)
& 0.6236 & 0.7083
& 0.6284 & 0.6623
& 0.5842 & 0.7452
& 0.5591 & 0.6317
& 0.6361 & 0.6000
& 0.4770 & 0.4929
& 0.6122 & 0.5000
& 0.5602 & 0.5400
& 0.4233 & 0.5450
& 0.5949 & 0.5312 \\

M3DM (Point-BERT)
& 0.7563 & 0.8889
& 0.8513 & 0.7984
& 0.6604 & \textbf{1.0000}
& 0.7081 & 0.9683
& 0.7863 & 0.8918
& 0.4933 & 0.6143
& 0.7523 & \textbf{1.0000}
& 0.6577 & 0.9167
& 0.7349 & 0.9083
& 0.6842 & 0.8375 \\

M3DM (Point-MAE)
& 0.8079 & 0.8851
& 0.8839 & 0.8475
& 0.6941 & \textbf{1.0000}
& 0.7207 & 0.9733
& 0.8057 & 0.9262
& 0.4946 & 0.6143
& 0.8510 & \textbf{1.0000}
& 0.6819 & 0.8550
& 0.7949 & 0.9271
& 0.7096 & 0.9266 \\

PatchCore (FPFH)
& 0.3878 & 0.6932
& 0.5324 & 0.5626
& 0.4258 & 0.9935
& 0.5040 & 0.4633
& 0.3179 & 0.5267
& 0.4794 & 0.5533
& 0.4923 & 0.6383
& 0.5999 & 0.6267
& 0.4806 & 0.4950
& 0.5328 & 0.5703 \\

PatchCore (Point-MAE)
& 0.7500 & 0.5391
& 0.6541 & 0.4410
& 0.6216 & 0.7061
& 0.4167 & 0.4867
& 0.4747 & 0.4967
& 0.3143 & 0.4911
& 0.6467 & 0.5000
& 0.6533 & 0.5233
& 0.4033 & 0.5233
& 0.4969 & 0.5219 \\

Reg3D-AD
& 0.6446 & 0.8258
& 0.7319 & 0.6967
& 0.6700 & 0.9937
& 0.5908 & 0.8950
& 0.7115 & 0.5033
& 0.4661 & 0.4875
& 0.7159 & 0.8700
& 0.5523 & 0.7200
& 0.6948 & 0.7300
& 0.6626 & 0.7969 \\

PO3AD
& 0.5458 & 0.4823
& 0.4538 & 0.4525
& 0.6573 & 0.1717
& 0.5792 & 0.6800
& 0.5525 & 0.5197
& 0.4140 & 0.4893
& 0.4639 & 0.5950
& 0.5201 & 0.5850
& 0.5275 & 0.4583
& 0.5121 & 0.4266 \\

PASDF
& 0.5977 & 0.6130
& 0.6219 & 0.7114
& \textbf{0.9441} & 0.3610
& 0.8274 & 0.9299
& 0.8761 & 0.8605
& 0.9860 & 0.9666
& 0.8484 & 0.8990
& 0.6586 & 0.6283
& 0.7720 & 0.6490
& 0.5037 & 0.7046 \\

Ours (SPA-SDF)
& \textbf{0.9472} & \textbf{0.9646}
& \textbf{0.8963} & \textbf{0.9623}
& 0.9228 & 0.8655
& \textbf{0.9219} & \textbf{0.9950}
& \textbf{0.9930} & \textbf{1.0000}
& \textbf{0.9963} & \textbf{1.0000}
& \textbf{0.9395} & 0.9830
& \textbf{0.8792} & \textbf{1.0000}
& \textbf{0.9030} & \textbf{0.9330}
& \textbf{0.7359} & \textbf{0.9469} \\
\midrule

\multirow{2}{*}{Method}
& \multicolumn{2}{c}{hinge2}
& \multicolumn{2}{c}{door2}
& \multicolumn{2}{c}{drawer2}
& \multicolumn{2}{c}{earphone\_case2}
& \multicolumn{2}{c}{fridge2}
& \multicolumn{2}{c}{injector2}
& \multicolumn{2}{c}{lamp2}
& \multicolumn{2}{c}{laptop2}
& \multicolumn{2}{c}{scissors2}
& \multicolumn{2}{c}{toilet\_seat2} \\
\cmidrule(lr){2-21}
& Pt & Obj & Pt & Obj & Pt & Obj & Pt & Obj & Pt & Obj
& Pt & Obj & Pt & Obj & Pt & Obj & Pt & Obj & Pt & Obj \\
\midrule
BTF (Raw)
& 0.5184 & 0.6190
& 0.5756 & 0.4670
& 0.4754 & 0.7217
& 0.5705 & 0.4300
& 0.5473 & 0.4710
& 0.5817 & 0.5367
& 0.5629 & 0.5300
& 0.5352 & 0.3500
& 0.5365 & 0.4267
& 0.6085 & 0.6317 \\

BTF (FPFH)
& 0.5778 & 0.9815
& 0.7179 & 0.5167
& 0.6260 & 0.5483
& 0.5668 & 0.4900
& 0.6704 & 0.6581
& 0.4864 & 0.4967
& 0.6454 & 0.6450
& 0.4944 & 0.2650
& 0.5997 & 0.4083
& 0.5474 & 0.3400 \\

M3DM (Point-BERT)
& 0.6627 & \textbf{1.0000}
& 0.7889 & 0.6333
& 0.6666 & 0.9750
& 0.6361 & 0.8933
& 0.8869 & 0.7167
& 0.8359 & 0.7233
& 0.7891 & \textbf{1.0000}
& 0.6255 & 0.7317
& 0.7613 & 0.8417
& 0.6997 & 0.8400 \\

M3DM (Point-MAE)
& \textbf{0.6692} & \textbf{1.0000}
& 0.9261 & 0.7833
& 0.6566 & \textbf{0.9990}
& 0.6885 & 0.9300
& 0.9209 & \textbf{0.7935}
& \textbf{0.8705} & \textbf{0.8017}
& 0.8338 & \textbf{1.0000}
& 0.6314 & 0.7383
& 0.8010 & \textbf{0.9600}
& \textbf{0.7718} & 0.8667 \\

PatchCore (FPFH)
& 0.4500 & \textbf{1.0000}
& 0.6886 & 0.4033
& 0.5681 & 0.5650
& 0.5101 & 0.6517
& 0.2670 & 0.5710
& 0.4466 & 0.6183
& 0.5137 & 0.5900
& 0.4602 & 0.6383
& 0.4286 & 0.5833
& 0.4577 & 0.5433 \\

PatchCore (Point-MAE)
& 0.4699 & 0.6586
& 0.4983 & 0.6017
& 0.5050 & 0.8567
& 0.5710 & 0.5750
& 0.4774 & 0.4967
& 0.4300 & 0.5733
& 0.4800 & 0.5783
& 0.3767 & 0.3783
& 0.6217 & 0.5400
& 0.5560 & 0.5267 \\

Reg3D-AD
& 0.6398 & 0.9792
& 0.7967 & 0.5750
& 0.6059 & 0.8433
& 0.6024 & 0.6600
& 0.7191 & 0.5597
& 0.8130 & 0.7317
& 0.7474 & 0.9767
& 0.5281 & 0.6367
& 0.6705 & 0.7217
& 0.6971 & 0.8083 \\

PO3AD
& 0.5056 & 0.5058
& 0.5532 & 0.4483
& 0.4711 & 0.1717
& 0.4571 & 0.4683
& 0.4876 & 0.4065
& 0.4519 & 0.4183
& 0.4990 & 0.3583
& 0.5120 & 0.6200
& 0.5054 & 0.4283
& 0.5308 & 0.3833 \\

PASDF
& 0.4980 & 0.5416
& 0.8346 & 0.5750
& \textbf{0.9494} & 0.6060
& 0.7059 & 0.6583
& 0.9193 & 0.6263
& 0.6228 & 0.6363
& 0.6702 & 0.6666
& 0.6794 & 0.5160
& 0.7484 & 0.4700
& 0.5767 & 0.5666 \\

Ours (SPA-SDF)
& 0.6012 & 0.9479
& \textbf{0.9820} & \textbf{0.9520}
& 0.9427 & 0.9050
& \textbf{0.8070} & \textbf{0.9717}
& \textbf{0.9862} & 0.7540
& 0.7730 & 0.4701
& \textbf{0.8692} & 0.9267
& \textbf{0.7622} & \textbf{0.8933}
& \textbf{0.9065} & 0.7330
& 0.7620 & \textbf{0.8830} \\
\midrule

\multirow{2}{*}{Method}
& \multicolumn{2}{c}{hinge3}
& \multicolumn{2}{c}{door3}
& \multicolumn{2}{c}{drawer3}
& \multicolumn{2}{c}{earphone\_case3}
& \multicolumn{2}{c}{fridge3}
& \multicolumn{2}{c}{injector3}
& \multicolumn{2}{c}{lamp3}
& \multicolumn{2}{c}{laptop3}
& \multicolumn{2}{c}{scissors3}
& \multicolumn{2}{c}{toilet\_seat3} \\
\cmidrule(lr){2-21}
& Pt & Obj & Pt & Obj & Pt & Obj & Pt & Obj & Pt & Obj
& Pt & Obj & Pt & Obj & Pt & Obj & Pt & Obj & Pt & Obj \\
\midrule
BTF (Raw)
& 0.5264 & 0.5799
& 0.6803 & 0.4850
& 0.4927 & 0.9567
& 0.4977 & 0.5017
& 0.4000 & 0.4633
& 0.5700 & 0.6567
& 0.5206 & 0.4950
& 0.5389 & 0.6417
& 0.5400 & 0.5317
& 0.5785 & 0.4267 \\

BTF (FPFH)
& 0.5543 & 0.7836
& 0.7372 & 0.5817
& 0.5002 & 0.9483
& 0.5743 & 0.3950
& 0.6835 & 0.4900
& 0.6504 & 0.5700
& 0.5644 & 0.5617
& 0.4935 & 0.7600
& 0.6236 & 0.5150
& 0.6817 & 0.4700 \\

M3DM (Point-BERT)
& 0.6236 & \textbf{1.0000}
& 0.9208 & 0.6917
& 0.5669 & \textbf{1.0000}
& 0.6154 & \textbf{0.9283}
& 0.8623 & 0.6700
& 0.7537 & 0.6867
& 0.8153 & 0.9350
& 0.5541 & 0.7900
& 0.7299 & 0.7467
& 0.7590 & 0.8233 \\

M3DM (Point-MAE)
& 0.6405 & \textbf{1.0000}
& \textbf{0.9542} & 0.6717
& 0.5867 & \textbf{1.0000}
& 0.6433 & 0.8633
& 0.9177 & 0.6250
& 0.8033 & \textbf{0.7017}
& 0.8899 & \textbf{0.9900}
& 0.5817 & 0.7150
& 0.7405 & \textbf{0.8867}
& 0.8343 & 0.8167 \\

PatchCore (FPFH)
& 0.3959 & 0.8333
& 0.6730 & 0.4300
& 0.4405 & \textbf{1.0000}
& 0.5299 & 0.4617
& 0.6353 & 0.4467
& 0.4572 & 0.4633
& 0.4936 & 0.6100
& 0.5527 & 0.6700
& 0.5272 & 0.5873
& 0.3441 & 0.5083 \\

PatchCore (Point-MAE)
& \textbf{0.7137} & 0.4971
& 0.4183 & 0.6367
& 0.3817 & 0.9670
& 0.5883 & 0.6167
& 0.6517 & 0.5400
& 0.2967 & 0.5267
& 0.4717 & 0.5833
& 0.5883 & 0.4967
& 0.3583 & 0.4617
& 0.5933 & 0.6650 \\

Reg3D-AD
& 0.6020 & 0.9190
& 0.8384 & 0.5600
& 0.5232 & 0.9183
& 0.5932 & 0.7550
& 0.7868 & 0.5233
& 0.7387 & 0.5167
& 0.7525 & 0.9317
& 0.5560 & 0.3767
& 0.6479 & 0.8167
& 0.6806 & 0.7783 \\

PO3AD
& 0.4867 & 0.0972
& 0.5458 & 0.5817
& 0.5262 & 0.5417
& 0.5010 & 0.5633
& 0.4662 & 0.4450
& 0.4700 & 0.4717
& 0.4597 & 0.4433
& 0.4975 & 0.4800
& 0.5963 & 0.3783
& 0.5512 & 0.3733 \\

PASDF
& 0.5420 & 0.5717
& 0.8899 & 0.5133
& \textbf{0.9780} & 0.5316
& 0.5624 & 0.5100
& 0.9525 & 0.6477
& \textbf{0.9625} & 0.6059
& 0.8178 & 0.7533
& 0.5405 & 0.5700
& 0.6055 & 0.5000
& 0.6580 & 0.6649 \\

Ours (SPA-SDF)
& 0.6960 & 0.9440
& 0.8650 & \textbf{0.9992}
& 0.8935 & 0.6083
& \textbf{0.6678} & 0.9170
& \textbf{0.9955} & \textbf{0.8770}
& 0.9207 & 0.4440
& \textbf{0.9018} & 0.9560
& \textbf{0.6929} & \textbf{0.8833}
& \textbf{0.8630} & 0.7650
& \textbf{0.9250} & \textbf{0.9567} \\
\midrule

\multirow{2}{*}{Method}
& \multicolumn{2}{c}{usb}
& \multicolumn{2}{c}{usb2}
& \multicolumn{2}{c}{usb3}
& \multicolumn{2}{c}{utility\_knife}
& \multicolumn{2}{c}{utility\_knife2}
& \multicolumn{2}{c}{utility\_knife3}
& \multicolumn{2}{c}{folding\_knife}
& \multicolumn{2}{c}{folding\_knife2}
& \multicolumn{2}{c}{folding\_knife3}
& \multicolumn{2}{c}{Mean} \\
\cmidrule(lr){2-21}
& Pt & Obj & Pt & Obj & Pt & Obj & Pt & Obj & Pt & Obj
& Pt & Obj & Pt & Obj & Pt & Obj & Pt & Obj & Pt & Obj \\
\midrule
BTF (Raw)
& 0.5091 & 0.4708
& 0.4930 & 0.4283
& 0.5142 & 0.4833
& 0.5463 & 0.5553
& 0.5311 & 0.6407
& 0.5978 & 0.5411
& 0.5433 & 0.4472
& 0.5844 & 0.7017
& 0.6271 & 0.6560
& 0.5379 & 0.5416 \\

BTF (FPFH)
& 0.5110 & 0.5799
& 0.5014 & 0.5433
& 0.5013 & 0.3200
& 0.5668 & 0.9683
& 0.5208 & 0.9508
& 0.4892 & 0.4594
& 0.6287 & 0.4736
& 0.5823 & 0.6467
& 0.6191 & 0.5733
& 0.5770 & 0.5893 \\

M3DM (Point-BERT)
& 0.6722 & 0.8670
& 0.5500 & 0.6633
& 0.5221 & 0.6667
& 0.6641 & \textbf{1.0000}
& 0.5699 & \textbf{1.0000}
& 0.6427 & 0.7938
& 0.8089 & \textbf{1.0000}
& 0.7084 & 0.9900
& 0.7055 & \textbf{1.0000}
& 0.7046 & 0.8572 \\

M3DM (Point-MAE)
& \textbf{0.7923} & \textbf{0.9630}
& 0.6007 & 0.7400
& 0.5801 & 0.6733
& 0.6693 & \textbf{1.0000}
& 0.6028 & \textbf{1.0000}
& 0.6862 & 0.7811
& 0.8576 & \textbf{1.0000}
& 0.7324 & \textbf{0.9983}
& 0.7385 & \textbf{1.0000}
& 0.7452 & 0.8782 \\

PatchCore (FPFH)
& 0.4416 & 0.6750
& 0.4677 & 0.5217
& 0.5004 & 0.5117
& 0.5174 & \textbf{1.0000}
& 0.4824 & \textbf{1.0000}
& 0.4436 & 0.4562
& 0.5654 & 0.5111
& 0.5732 & 0.6971
& 0.4555 & 0.6500
& 0.5019 & 0.6397 \\

PatchCore (Point-MAE)
& 0.4615 & 0.5042
& 0.4300 & 0.4283
& 0.5883 & 0.3933
& 0.6167 & 0.9850
& 0.6475 & 0.8240
& 0.4436 & 0.4562
& 0.4389 & 0.4250
& 0.3083 & 0.3367
& 0.5855 & 0.4443
& 0.5043 & 0.5892 \\

Reg3D-AD
& 0.6423 & 0.7542
& 0.4950 & 0.5217
& 0.5419 & 0.7167
& 0.6542 & 0.9433
& 0.5227 & \textbf{1.0000}
& 0.5942 & 0.6047
& 0.6965 & 0.8014
& 0.6255 & 0.8567
& 0.6048 & 0.8617
& 0.6502 & 0.7479 \\

PO3AD
& 0.5327 & 0.3229
& 0.4887 & 0.4233
& 0.5412 & 0.4300
& 0.5076 & 0.3533
& 0.5217 & 0.3695
& 0.4749 & 0.6234
& 0.5297 & 0.4986
& 0.4015 & 0.3333
& 0.3929 & 0.3550
& 0.5035 & 0.4453 \\

PASDF
& 0.5110 & 0.5799
& 0.6540 & 0.5716
& 0.7509 & 0.4400
& 0.8050 & 0.6670
& 0.6724 & 0.7150
& 0.7849 & 0.4531
& 0.6838 & 0.6160
& 0.6964 & 0.5716
& 0.6838 & 0.6160
& 0.6620 & 0.6840 \\

Ours (SPA-SDF)
& \textbf{0.7923} & \textbf{0.9630}
& \textbf{0.8276} & \textbf{0.7617}
& \textbf{0.8988} & \textbf{0.7967}
& \textbf{0.9728} & 0.9781
& \textbf{0.7379} & 0.8831
& \textbf{0.8390} & \textbf{0.8578}
& \textbf{0.9150} & 0.9560
& \textbf{0.8947} & 0.9000
& \textbf{0.9150} & 0.9560
& \textbf{0.8651} & \textbf{0.8843} \\
\bottomrule
\end{tabular}
\end{adjustbox}\vspace{0mm}
\end{table*}

\paragraph{SPA-SDF benefits from replacing the rigid prior.}
\textcolor{black}{By modeling normality as a continuous pose-conditioned manifold,
SPA-SDF outperforms all baselines by a clear margin, achieving 0.884
and 0.874 Obj-AUROC on seen and unseen splits respectively. The small
seen/unseen gap of 0.010 demonstrates that the Fourier-encoded
manifold generalizes smoothly to novel joint configurations rather
than memorizing training poses, confirming that breaking the rigid
prior enables genuine articulation generalization.}

\subsection{Ablation Study}
\label{sec:ablation}

\paragraph{Component ablation.}
\textcolor{black}{Table~\ref{tab:ablation_spasdf} evaluates each model component by
progressive inclusion. The decisive contribution is the articulation
embedding (A4), which raises Seen-Obj from 0.707 to 0.852, a gain of
14.5 points, directly quantifying the benefit of breaking the rigid
prior with explicit pose conditioning. The full SPA-SDF model provides
additional improvement on the unseen split (10.4 points over A4 in
Unseen-Obj), demonstrating that part-consistent regularization
supports generalization to novel articulation states.}

\paragraph{Articulation encoding ablation.}
\textcolor{black}{Table~\ref{tab:pose_ablation} examines the effect of encoding strategy.
Removing articulation entirely (A0) establishes the performance
ceiling of the rigid prior, which is substantially below all
pose-conditioned variants. Multi-frequency Fourier encoding
(A3 to A4) provides consistent improvements over raw $\psi$ (A1),
and the full model (A5: Fourier $L{=}16$ with Pose MLP) achieves the
best results with the smallest seen/unseen AUROC gap, indicating that
richer articulation encoding improves generalization beyond the
training pose distribution.}

\subsection{Qualitative Results}

\begin{table}[t!]
\caption{Component ablation study of SPA-SDF.
Mean point-level (Pt) and object-level (Obj) AUROC on seen and unseen categories.} \vspace{-2mm}
\label{tab:ablation_spasdf}
\centering
\small
\setlength{\tabcolsep}{4pt}
\begin{tabular}{lcccc}
\toprule
\textbf{Method} 
& \textbf{Seen-Pt} 
& \textbf{Seen-Obj} 
& \textbf{Unseen-Pt} 
& \textbf{Unseen-Obj} \\
\midrule

A0 (xyz)              & 0.5981 & 0.6335 & 0.6392 & 0.6664 \\
A1 (xyz+PE)           & 0.6621 & 0.6847 & 0.6952 & 0.6891 \\
A2 (xyz+shape)        & 0.6393 & 0.6759 & 0.6663 & 0.6930 \\
A3 (xyz+PE+shape)     & 0.6889 & 0.7070 & 0.7140 & 0.7150 \\
A4 (xyz+PE+pose)      & 0.8302 & 0.8520 & 0.8151 & 0.7701 \\

\midrule
\textbf{Full SPA-SDF} & \textbf{0.8651} & \textbf{0.8843} & \textbf{0.8410} & \textbf{0.8740} \\
\bottomrule
\end{tabular}\vspace{5mm}
\end{table}

\begin{figure}[t]
    \centering
    \includegraphics[width=\columnwidth]{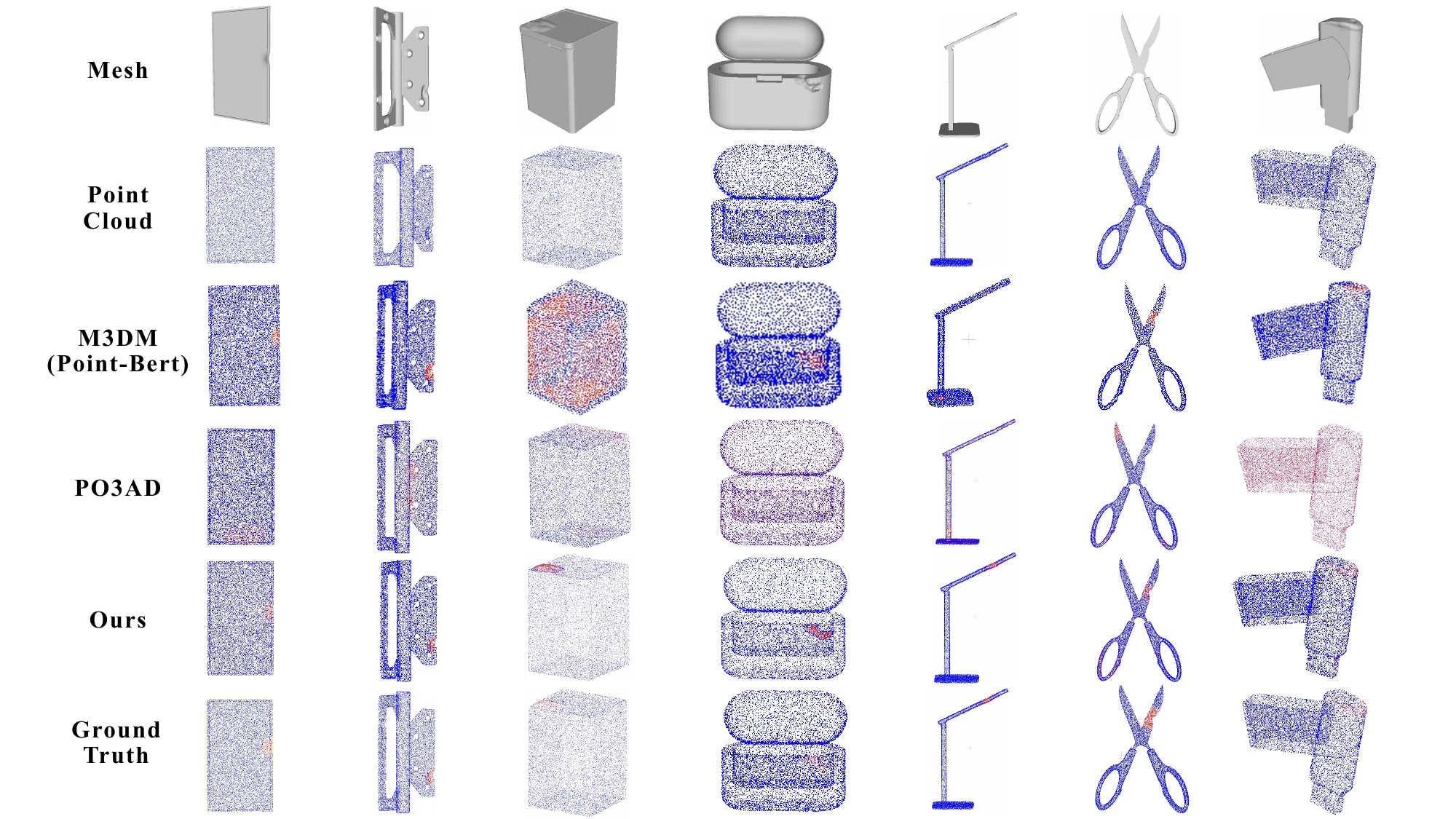}
    \caption{Qualitative comparison of anomaly localization under articulated pose variations.}
    \Description{
    Qualitative comparison of anomaly localization results across different methods on articulated objects. 
    The figure shows input meshes, predicted anomaly maps, and ground truth labels. 
    Prior methods produce false positives under articulation, while SPA-SDF provides more accurate and coherent localization.
    }
    \label{fig:compare}\vspace{3mm}
\end{figure}

Qualitative comparisons with existing methods are shown in Fig.~\ref{fig:compare}.
Under articulated pose variations, prior methods exhibit distinct failure modes.
M3DM is able to roughly localize defect regions, but produces noisy responses that spread across normal areas, revealing its sensitivity to pose-induced geometric variations.
In contrast, PO3AD largely fails to identify defect locations, often yielding overly smooth or near-uniform predictions with weak anomaly responses.

Our method, by contrast, produces more precise and spatially coherent anomaly localization that closely aligns with ground truth.
By explicitly modeling articulation, SPA-SDF effectively separates pose-induced geometric variations from true structural defects, significantly reducing false positives while preserving accurate defect regions across diverse articulated configurations.

\begin{figure}[t]
    \centering
    \includegraphics[width=\columnwidth]{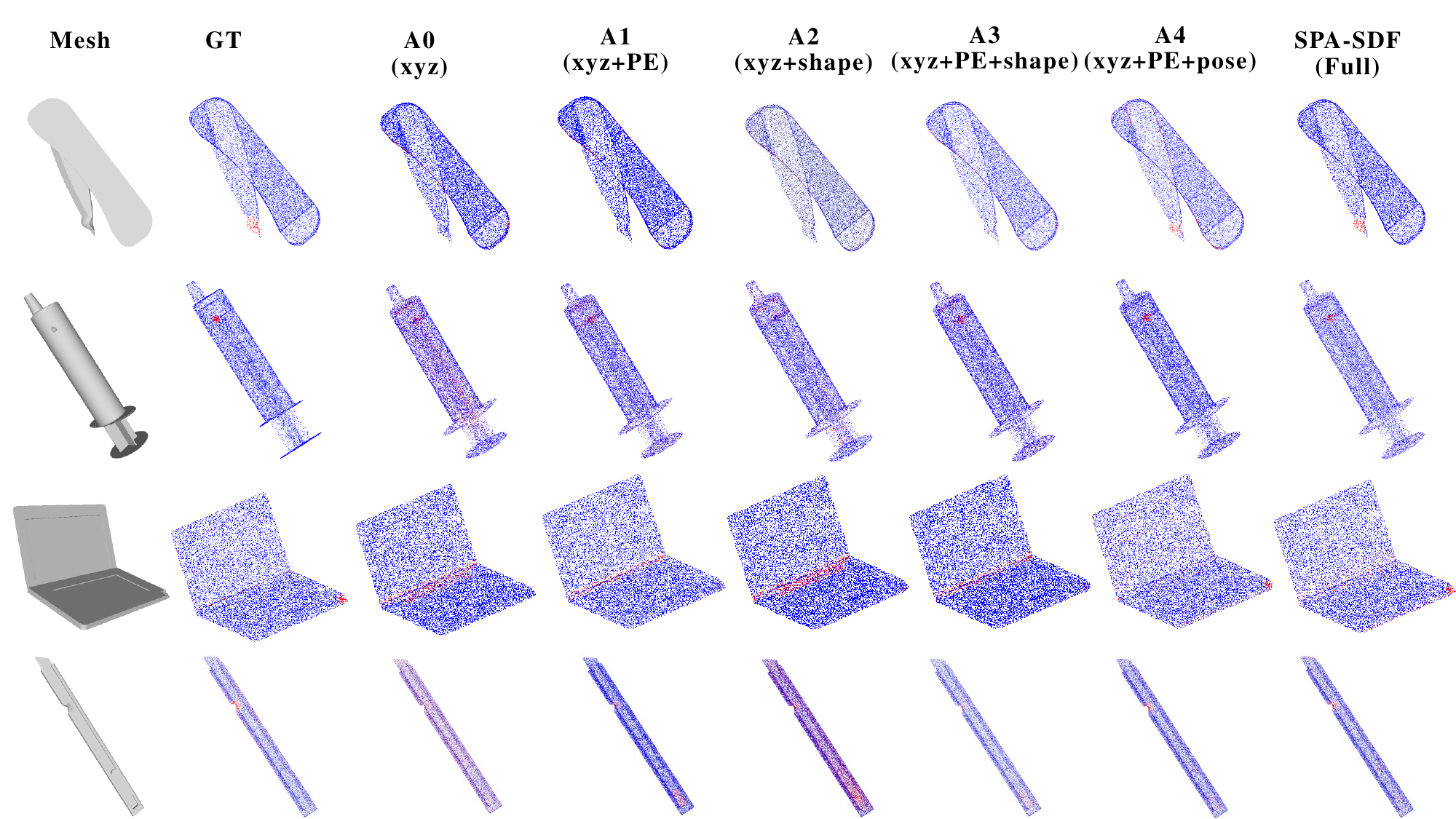}
    \caption{
    Qualitative comparison of component effectiveness in SPA-SDF.
    From left to right: input mesh, ground truth, and ablation variants including 
    A0 (xyz), A1 (xyz+PE), A2 (xyz+shape), A3 (xyz+PE+shape), A4 (xyz+PE+pose), and the full SPA-SDF model.
    }
    \label{fig:qualitative_ablation}\vspace{5mm}
\end{figure}

We further visualize the impact of each component in SPA-SDF in Fig.~\ref{fig:qualitative_ablation}.
Without pose conditioning (A0--A2), the model tends to assign high anomaly scores to normal regions undergoing articulation, reflecting a strong bias toward rigid assumptions.
In particular, articulation-induced geometric variations are frequently misinterpreted as structural anomalies.
Adding positional encoding (A1) slightly improves geometric representation but does not resolve the ambiguity between pose-induced deformation and true defects.
Introducing the shape prior (A2 and A3) enhances structural consistency, yet remains insufficient to fully disentangle pose-dependent variations, especially under large articulation changes.

In contrast, incorporating pose conditioning (A4) significantly reduces false positives in articulated regions, highlighting the necessity of explicitly modeling articulation.
The full model further achieves effective disentanglement of pose and structure, producing sharper and more accurate anomaly localization that aligns well with ground truth, particularly in challenging articulated scenarios.

\begin{table}[t]
    \centering
    \caption{Ablation study on pose encoding in SPA-SDF. 
    We report mean point-level (Pt) and object-level (Obj) AUROC over all 39 categories for both seen and unseen splits.} \vspace{-2mm}
    \label{tab:pose_ablation}
    \setlength{\tabcolsep}{4pt}
    \small
    \begin{tabular}{lcccc}
    \toprule
    \textbf{Method} 
    & \textbf{Seen-Pt} 
    & \textbf{Seen-Obj} 
    & \textbf{Unseen-Pt} 
    & \textbf{Unseen-Obj} \\
    \midrule
    
    A0 (w/o $\theta$)            & 0.7196 & 0.6348 & 0.7373 & 0.6095 \\
    A1 (raw $\theta$)            & 0.8338 & 0.8500 & 0.8301 & 0.8450 \\
    A2 (fixed $\sin/\cos$)       & 0.8356 & 0.8751 & 0.8350 & 0.8563 \\
    A3 (Fourier $L{=}8$)         & 0.8398 & 0.8781 & 0.8212 & 0.8175 \\
    A4 (Fourier $L{=}16$)        & 0.8431 & 0.8679 & 0.8324 & 0.8207 \\
    
    \midrule
    \textbf{SPA-SDF (full)} 
    & \textbf{0.8651} 
    & \textbf{0.8843} 
    & \textbf{0.8410} 
    & \textbf{0.8740} \\
    
    \bottomrule
    \end{tabular}\vspace{7mm}
\end{table}

\section{Conclusion}
\label{sec:conclusion}

\textcolor{black}{We presented ArtiAD, the first large-scale benchmark for articulated 3D anomaly detection, together with SPA-SDF, a pose-conditioned implicit baseline that replaces the rigid prior with a continuous pose-dependent normality manifold. The central finding is that the rigid prior, which has served as an unexamined foundation of the field, breaks systematically for objects with hinge and sliding joints. Comprehensive experiments across 39 categories confirm that all methods enforcing this prior degrade substantially under articulation variation, while SPA-SDF achieves significantly stronger performance by modeling normality as a function of joint configuration rather than as a fixed canonical geometry. We hope this work motivates the community to move beyond the rigid prior and develop anomaly detection methods that embrace, rather than suppress, the kinematic structure of
real-world objects.}

\paragraph{Limitations and Future Work.}
\textcolor{black}{ArtiAD currently covers single degree-of-freedom (DoF) mechanisms, and SPA-SDF assumes a known kinematic type per category with articulation estimation performed by exhaustive grid search, which scales poorly beyond single-DoF settings. Extending the benchmark to multi-DoF articulated structures and developing differentiable articulation estimation within the implicit field are natural next steps. Incorporating real-scanned objects alongside synthetic data would improve ecological validity. More broadly, designing methods that jointly infer articulation structure and detect defects in a unified framework, rather than treating pose estimation as a preprocessing step, represents a compelling direction for future work.}

\bibliographystyle{ACM-Reference-Format}
\bibliography{cite}
\appendix

\setcounter{table}{0}
\setcounter{figure}{0}
\renewcommand{\thetable}{\Alph{table}}
\renewcommand{\thefigure}{\Alph{figure}}

\section{ArtiAD Dataset Details}
\subsection{Articulation Space Coverage}

ArtiAD explicitly models articulation as a continuous variable.
For hinge-based objects, the articulation parameter $\theta$ is sampled
from a bounded angular range (e.g., $[0^\circ, 120^\circ]$), while
for prismatic objects, the translation parameter $d$ is sampled from
a valid displacement interval.

\paragraph{Physically Plausible Articulation.}
To ensure realism, all articulation variations strictly follow
the underlying kinematic constraints of each object.
Specifically, joint motion is defined on the basis of the physical structure
of the object, including hinge axes, rotation limits, and sliding directions.
The articulation ranges are manually specified to match
real-world operational limits, avoiding physically invalid configurations
(e.g., self-intersections or disjoint parts).

For hinge-based objects, rotation is constrained to a single axis with
bounded angles reflecting realistic usage (e.g., door opening angles).
For prismatic objects, translation is restricted to feasible sliding
ranges along predefined directions.
This design ensures that all generated shapes correspond to
physically valid configurations rather than arbitrary geometric deformations.

\paragraph{Continuous and Dense Sampling.}
To ensure sufficient coverage of the articulation manifold,
we densely sample articulation states using uniform sampling
over the valid parameter space.
This results in a continuous distribution of geometrically valid shapes,
enabling models to learn smooth pose-dependent variations.

\paragraph{Category-Specific Parameterization.}
Each object category is parameterized independently according to its
kinematic structure, ensuring that articulation sampling is tailored
to the specific motion pattern of the object rather than using a
shared generic configuration.

Table~\ref{tab:articulation_details} illustrates the distribution of the articulation parameters
for all the object categories in ArtiAD.

\begin{table}[t]
\centering
\caption{
Articulation parameter ranges for all 39 categories in ArtiAD.
Rotational joints are parameterized by angle $\theta$ (in degrees),
while prismatic joints are parameterized by normalized displacement $d$.
}\vspace{-2mm}
\label{tab:articulation_details}
\small
\setlength{\tabcolsep}{5pt}

\begin{tabular}{lcc|lcc}
\toprule
\textbf{Category} & \textbf{$\psi$} & \textbf{Range}
& \textbf{Category} & \textbf{$\psi$} & \textbf{Range} \\
\midrule

Hinge   & $\theta$ & $[0^\circ,359^\circ]$
& Door   & $\theta$ & $[0^\circ,90^\circ]$ \\
Hinge2  & $\theta$ & $[0^\circ,180^\circ]$
& Door2  & $\theta$ & $[0^\circ,150^\circ]$ \\
Hinge3  & $\theta$ & $[0^\circ,180^\circ]$
& Door3  & $\theta$ & $[0^\circ,150^\circ]$ \\

Drawer  & $d$ & $[0,85]$
& Earphone  & $\theta$ & $[0^\circ,135^\circ]$ \\
Drawer2 & $d$ & $[0,85]$
& Earphone2 & $\theta$ & $[0^\circ,135^\circ]$ \\
Drawer3 & $d$ & $[0,250]$
& Earphone3 & $\theta$ & $[0^\circ,135^\circ]$ \\

Knife   & $\theta$ & $[0^\circ,180^\circ]$
& Fridge   & $\theta$ & $[0^\circ,90^\circ]$ \\
Knife2  & $\theta$ & $[0^\circ,180^\circ]$
& Fridge2  & $\theta$ & $[0^\circ,120^\circ]$ \\
Knife3  & $\theta$ & $[0^\circ,180^\circ]$
& Fridge3  & $\theta$ & $[0^\circ,150^\circ]$ \\

Injector  & $d$ & $[0,180]$
& Lamp   & $\theta$ & $[0^\circ,180^\circ]$ \\
Injector2 & $d$ & $[0,355]$
& Lamp2  & $\theta$ & $[0^\circ,180^\circ]$ \\
Injector3 & $d$ & $[0,345]$
& Lamp3  & $\theta$ & $[0^\circ,180^\circ]$ \\

Laptop  & $\theta$ & $[0^\circ,120^\circ]$
& Scissors  & $\theta$ & $[10^\circ,120^\circ]$ \\
Laptop2 & $\theta$ & $[0^\circ,120^\circ]$
& Scissors2 & $\theta$ & $[10^\circ,120^\circ]$ \\
Laptop3 & $\theta$ & $[0^\circ,120^\circ]$
& Scissors3 & $\theta$ & $[10^\circ,120^\circ]$ \\

Seat   & $\theta$ & $[0^\circ,100^\circ]$
& USB   & $\theta$ & $[0^\circ,359^\circ]$ \\
Seat2  & $\theta$ & $[0^\circ,120^\circ]$
& USB2  & $\theta$ & $[0^\circ,359^\circ]$ \\
Seat3  & $\theta$ & $[0^\circ,120^\circ]$
& USB3  & $\theta$ & $[0^\circ,359^\circ]$ \\

UtilKnife  & $d$ & $[0,90]$
& UtilKnife2 & $d$ & $[0,100]$ \\
UtilKnife3 & $d$ & $[0,160]$
& \multicolumn{3}{c}{--} \\
\bottomrule
\end{tabular}
\vspace{2mm}
\end{table}

\subsection{Normal Sample Generation and SDF Preparation}

Normal samples are generated through a multi-stage preprocessing pipeline that produces cleaned articulated meshes, part-aware point clouds, and signed-distance supervision.

\paragraph{Boolean-Based Mesh Cleaning.} 
For each articulation state, the object is first represented by multiple part variants (e.g., \texttt{part0}, \texttt{part1}). Before point cloud generation, Boolean operations are applied to remove overlapping internal faces between adjacent parts. This step eliminates redundant interior surfaces caused by part intersections, so that the final articulated mesh contains only physically visible geometry.

\paragraph{Part-Aware Point Labeling.} 
After Boolean processing, dense surface points are sampled from the merged articulated mesh. To assign a part label to each sampled point, we additionally load the corresponding part meshes under the same articulation state and assign each point to the nearest part. The resulting point cloud is stored together with its point-wise \texttt{part-id}, which provides explicit part-level supervision.

\paragraph{Category-Level Normalization.} 
To ensure consistency across different articulation states of the same category, we compute category-level normalization parameters from all training samples. Specifically, we load all point clouds in the training split of one category, aggregate their vertices, and compute a global center as the mean of all vertices:
\begin{equation}
\mathbf{c} = \frac{1}{N} \sum_{i=1}^{N} \mathbf{x}_i,
\end{equation}
where $\mathbf{x}_i \in \mathbb{R}^3$ denotes a vertex from the training data. We then subtract this global center and define the global scale $s$ as the maximum absolute coordinate value after centering:
\begin{equation}
s = \max_{i} \| \mathbf{x}_i - \mathbf{c} \|_{\infty}.
\end{equation}
The normalization operation is defined as:
\begin{equation}
\mathbf{x}' = \frac{\mathbf{x} - \mathbf{c}}{s}.
\end{equation}
The normalization parameters, including the category-level center and scale, are stored per category to ensure a shared coordinate system. This design guarantees that all articulation states are normalized within a consistent spatial reference frame, rather than being processed independently.

\paragraph{Signed Distance Field (SDF) Sampling.} 
Based on the normalized point cloud and the corresponding Boolean mesh, we generate the final SDF supervision file for each sample. The implementation uses four types of query points: (i) on-surface points, obtained by subsampling the surface point cloud; (ii) near-surface points, generated by adding Gaussian perturbations to surface samples; (iii) points uniformly sampled inside the sample bounding box; and (iv) points uniformly sampled in the normalized global volume. For off-surface query points, part labels are propagated from the nearest surface point via nearest-neighbor search.

\paragraph{Signed Distance Computation.} 
Signed distances are computed against the triangle mesh after applying the same category-level normalization used for the point cloud. In our implementation, the Boolean mesh is loaded as a triangle mesh, cleaned by removing unreferenced vertices and degenerate or duplicate faces, and then used to evaluate signed distance values for all sampled query points.

\paragraph{Final Training File Format.} 
For each articulation state, the final output is stored as an \texttt{npz} file containing data of the form:
\begin{equation}
(\mathbf{x}, \mathrm{sdf}, \mathrm{part\_id}),
\end{equation}
where $\mathbf{x} \in \mathbb{R}^3$ denotes the normalized 3D query location, $\mathrm{sdf} \in \mathbb{R}$ is the signed distance to the articulated mesh, and $\mathrm{part\_id} \in \mathbb{Z}$ denotes the propagated part label.

\subsection{Structural Anomaly Generation and Abnormal Sample Extraction}

We simulate six types of structural anomalies, including
\textit{dents}, \textit{bulges}, \textit{fractures},
\textit{bending deformations}, \textit{surface distortions},
and \textit{missing material}, following the definitions in the main paper.

\paragraph{Blender-Based Anomaly Injection.}
All anomalies are created in Blender via manual mesh editing.
We introduce localized geometric perturbations on articulated object meshes
to simulate realistic structural defects while maintaining overall structural consistency.
For each anomaly instance, we control the spatial location,
deformation scale, and deformation intensity,
thereby producing diverse anomaly patterns across categories and articulation states.

\paragraph{Decoupling Articulation and Anomaly.}
Anomaly injection is performed
\emph{after} articulation transformation.
This ensures that pose-induced and defect-induced geometry coexist
but originate from independent generative factors.
As a result, geometric deviations introduced by articulation remain
distinct from those caused by structural anomalies.

\paragraph{Abnormal Surface Sampling.}
For each anomalous sample, we first load the defect mesh and its paired
normal mesh under the same articulation state.
We then uniformly sample
\begin{equation}
N_{\mathrm{dense}} = 100{,}000
\end{equation}
surface points from the anomalous mesh to obtain a dense abnormal point cloud.

\paragraph{Part-Aware Labeling.}
To assign part labels to abnormal points, we load the corresponding part meshes.
For each sampled point $\mathbf{x}$, we compute its closest distance to the mesh of each part
and assign the part identity according to the nearest one.
This yields a point-wise \texttt{part-id} annotation for all abnormal samples.

\paragraph{Point-Level Ground Truth Generation.}
Point-level anomaly labels are generated by comparing each sampled abnormal point
with the paired normal mesh.
Specifically, for each abnormal point $\mathbf{x}$, we compute its distance to
the closest point on the normal mesh.
A point is labeled as anomalous if this distance exceeds a threshold
\begin{equation}
\tau = \alpha \cdot \left\| \mathbf{b}_{\max} - \mathbf{b}_{\min} \right\|_2,
\end{equation}
where $\mathbf{b}_{\max}$ and $\mathbf{b}_{\min}$ denote the bounding-box corners
of the normal mesh, and $\alpha = 0.001$.

The anomaly label is defined as
\begin{equation}
g(\mathbf{x}) = 
\begin{cases} 
1, & \text{if } d(\mathbf{x}, \mathcal{M}_{\text{normal}}) > \tau, \\
0, & \text{otherwise.}
\end{cases}
\end{equation}

\paragraph{Dense-to-Sparse Sampling.}
After obtaining the dense abnormal point cloud and its labels,
we randomly shuffle the points and uniformly select
\begin{equation}
N_{\mathrm{final}} = 16{,}384
\end{equation}
points without replacement to form the final abnormal point cloud.

\paragraph{Category-Level Normalization.}
The sampled abnormal points are normalized using the same category-level
normalization parameters as normal samples.
Given the category-level center $\mathbf{c}$ and scale $s$,
the normalized point is computed as
\begin{equation}
\mathbf{x}' = \frac{\mathbf{x} - \mathbf{c}}{s}.
\end{equation}
This ensures that both normal and abnormal samples are represented
in a consistent coordinate system.

\paragraph{Saved Outputs.}
For each abnormal sample, we store:
(i) the dense abnormal point cloud with \texttt{part-id},
(ii) the corresponding dense anomaly labels,
(iii) the normalized point cloud with 16{,}384 points and \texttt{part-id}, and
(iv) the corresponding anomaly labels.

The abnormal point cloud is represented as
$(\mathbf{x}, \mathrm{part\text{-}id})$,
and the ground-truth annotation is represented as
$(\mathbf{x}, \mathrm{part\text{-}id}, g(\mathbf{x}))$.

\begin{figure*}[t]
\centering

\begin{subfigure}{\textwidth}
\centering
\includegraphics[width=\textwidth, trim=0 40 0 40, clip]{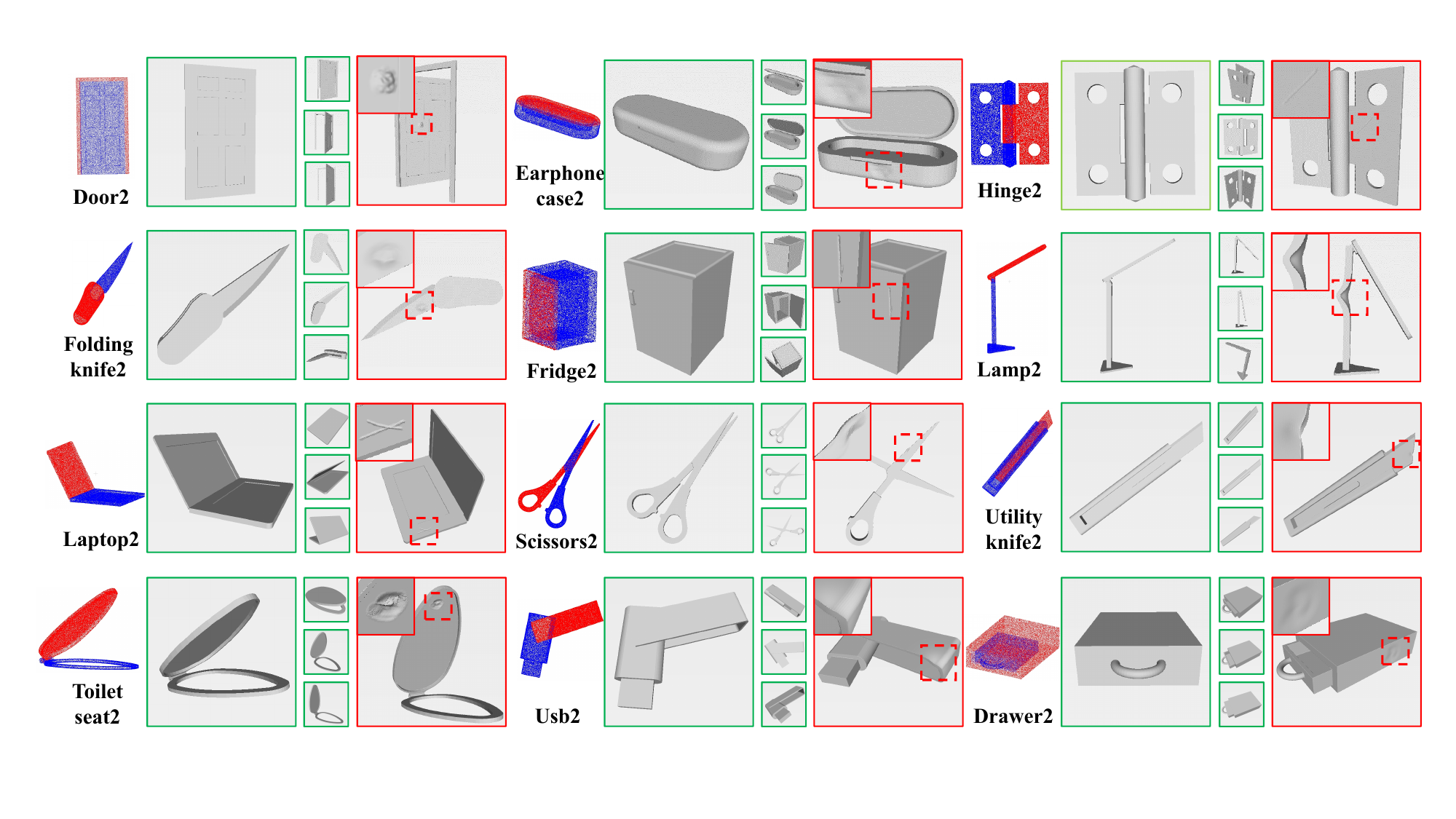}
\caption{Variant 2}
\end{subfigure}

\vspace{3mm}

\begin{subfigure}{\textwidth}
\centering
\includegraphics[width=\textwidth, trim=0 40 0 40, clip]{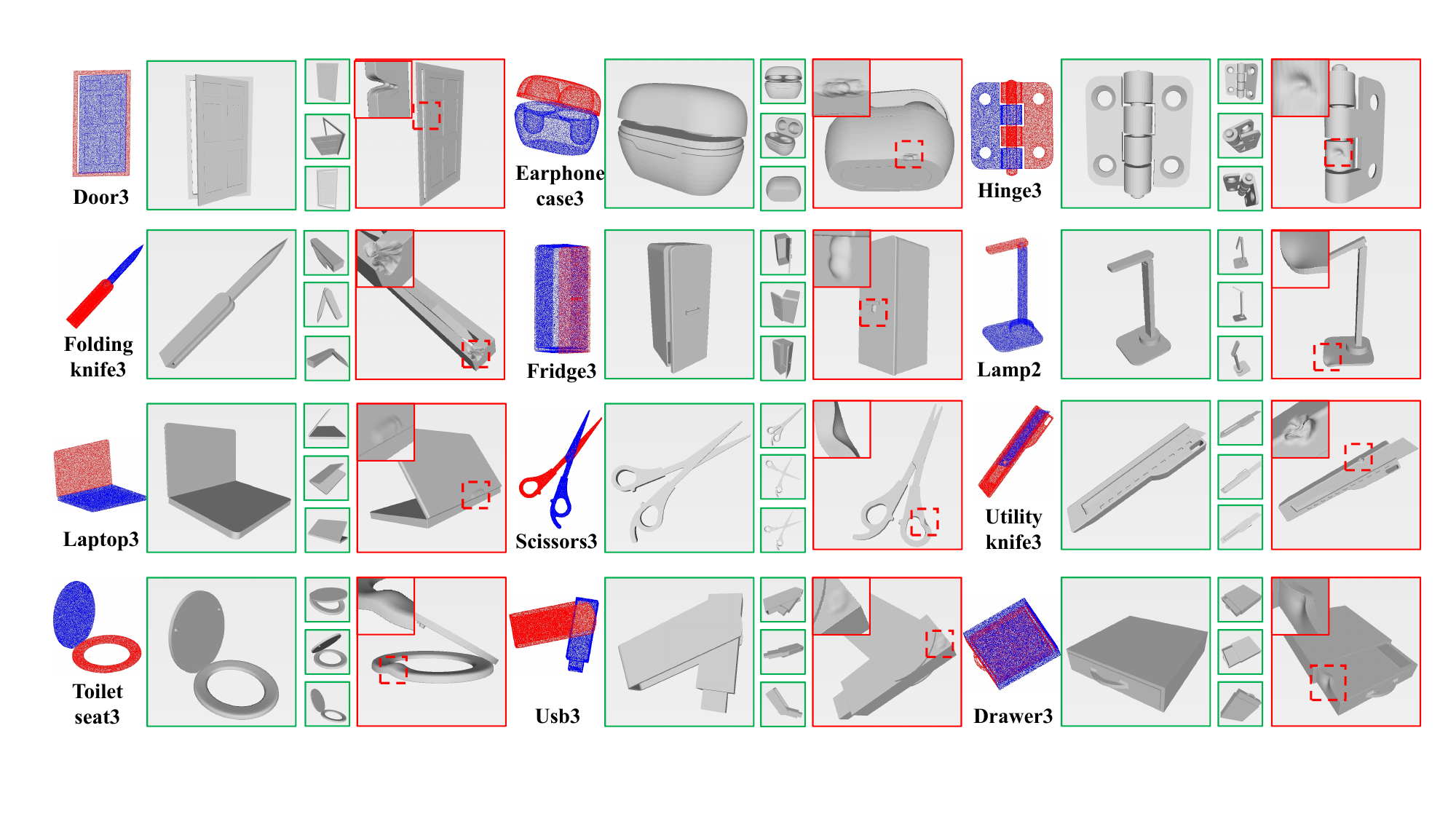}
\caption{Variant 3}
\end{subfigure}

\vspace{-3mm}

\caption{Additional qualitative examples of ArtiAD.}
\label{fig:anomaly_variant23}
\end{figure*}

\paragraph{Additional Visualization.}
The main paper presents representative anomaly examples for the first geometric variant of each category.
Here, we provide qualitative examples for the remaining two variants
(i.e., variant 2 and variant 3) in Fig.~\ref{fig:anomaly_variant23}.
These examples demonstrate that the anomaly generation and extraction process
remains consistent across different geometric variants while preserving
articulation-dependent geometry.

\section{Additional Experimental Results}

\subsection{Results under Unseen Articulation Settings}

In the main paper, we report overall performance and results under the Seen setting for clarity and conciseness.
Here, we provide additional results under the Unseen setting, which evaluates the model's ability to generalize to articulation states outside the training range.

Table~\ref{tab:benchmark_unseen} presents the detailed results for all methods in the Unseen setting.
Compared to the Seen setting, this scenario is more challenging due to the presence of articulation configurations that are not observed during training.

Our method consistently achieves superior performance under the Unseen setting, demonstrating strong generalization capability across diverse articulation states.

\begin{table*}[t]
\centering
\caption{Comparison on unseen classes. We report point-level AUROC (Pt) and object-level AUROC (Obj).}
\label{tab:benchmark_unseen}
\footnotesize
\setlength{\tabcolsep}{1.5pt}
\begin{adjustbox}{width=\textwidth}
\begin{tabular}{l*{10}{cc}}
\toprule
\multirow{2}{*}{Method}
& \multicolumn{2}{c}{hinge}
& \multicolumn{2}{c}{door}
& \multicolumn{2}{c}{drawer}
& \multicolumn{2}{c}{earphone\_case}
& \multicolumn{2}{c}{fridge}
& \multicolumn{2}{c}{injector}
& \multicolumn{2}{c}{lamp}
& \multicolumn{2}{c}{laptop}
& \multicolumn{2}{c}{scissors}
& \multicolumn{2}{c}{toilet\_seat} \\
\cmidrule(lr){2-21}
& Pt & Obj & Pt & Obj & Pt & Obj & Pt & Obj & Pt & Obj
& Pt & Obj & Pt & Obj & Pt & Obj & Pt & Obj & Pt & Obj \\
\midrule
BTF (Raw)
& 0.4936 & 0.5650
& 0.5593 & 0.5722
& 0.5367 & 0.8849
& 0.5084 & 0.3083
& 0.4863 & 0.4161
& 0.5538 & 0.4929
& 0.5770 & 0.5650
& 0.5068 & 0.4290
& 0.5986 & 0.4583
& 0.5718 & 0.5382 \\

BTF (FPFH)
& 0.6214 & 0.6900
& 0.5730 & 0.3944
& 0.6116 & 0.6163
& 0.5700 & 0.4450
& 0.6019 & 0.5484
& 0.4883 & 0.5857
& 0.6339 & 0.5800
& 0.5363 & 0.4710
& 0.5964 & 0.4700
& 0.5242 & 0.7250 \\

M3DM (Point-BERT)
& 0.7747 & 0.9013
& 0.7164 & 0.9111
& 0.6209 & \textbf{1.0000}
& 0.6346 & 0.9033
& 0.8875 & 0.9484
& 0.5486 & 0.5790
& 0.8461 & 0.9967
& 0.6128 & 0.9161
& 0.7583 & 0.7083
& 0.6376 & \textbf{0.9235} \\

M3DM (Point-MAE)
& 0.8347 & \textbf{0.9650}
& 0.8013 & \textbf{0.9519}
& 0.6596 & \textbf{1.0000}
& 0.6701 & \textbf{0.9433}
& 0.9499 & 0.9355
& 0.5261 & 0.4036
& 0.8917 & \textbf{1.0000}
& 0.6267 & 0.8661
& 0.8101 & 0.8067
& 0.6792 & 0.9147 \\

PatchCore (FPFH)
& 0.3857 & 0.4337
& 0.6031 & 0.4630
& 0.4326 & 0.9965
& 0.5078 & 0.5817
& 0.3941 & 0.5855
& 0.4755 & 0.5536
& 0.4869 & 0.5200
& 0.5398 & 0.6919
& 0.4321 & 0.4533
& 0.5301 & 0.5103 \\

PatchCore (Point-MAE)
& 0.4775 & 0.6787
& 0.3463 & 0.4074
& 0.4326 & 0.7151
& 0.3983 & 0.5083
& 0.4323 & 0.5016
& 0.4625 & 0.5071
& 0.3233 & 0.5950
& 0.5258 & 0.4677
& 0.3900 & 0.5150
& 0.6618 & 0.5147 \\

Reg3D-AD
& 0.6856 & 0.7250
& 0.6837 & 0.7352
& 0.6565 & 0.9826
& 0.5692 & 0.7233
& 0.7668 & 0.8339
& 0.4593 & 0.6357
& 0.8033 & 0.9950
& 0.5748 & 0.6468
& 0.7400 & 0.9300
& 0.6326 & 0.7485 \\

PO3AD
& 0.5291 & 0.5950
& 0.4564 & 0.5444
& 0.6157 & 0.2360
& 0.5339 & 0.6633
& 0.5158 & 0.4790
& 0.3890 & 0.4571
& 0.4515 & 0.5050
& 0.4984 & 0.5435
& 0.5107 & 0.4483
& 0.4850 & 0.4294 \\

PASDF
& 0.5660 & 0.5850
& 0.6530 & 0.6330
& \textbf{0.9450} & 0.3950
& 0.7950 & 0.9090
& 0.8630 & 0.6290
& \textbf{0.9050} & \textbf{0.9530}
& 0.8530 & 0.8160
& 0.7040 & 0.6258
& 0.6776 & 0.6950
& 0.5310 & 0.6647 \\

Ours (SPA-SDF)
& \textbf{0.9599} & 0.9388
& \textbf{0.8489} & 0.9444
& 0.9396 & 0.8600
& \textbf{0.8749} & 0.9317
& \textbf{0.9911} & \textbf{0.9887}
& 0.6000 & 0.6304
& \textbf{0.8959} & 0.9800
& \textbf{0.7535} & \textbf{0.9226}
& \textbf{0.8844} & \textbf{0.9600}
& \textbf{0.7091} & 0.9132 \\
\midrule

\multirow{2}{*}{Method}
& \multicolumn{2}{c}{hinge2}
& \multicolumn{2}{c}{door2}
& \multicolumn{2}{c}{drawer2}
& \multicolumn{2}{c}{earphone\_case2}
& \multicolumn{2}{c}{fridge2}
& \multicolumn{2}{c}{injector2}
& \multicolumn{2}{c}{lamp2}
& \multicolumn{2}{c}{laptop2}
& \multicolumn{2}{c}{scissors2}
& \multicolumn{2}{c}{toilet\_seat2} \\
\cmidrule(lr){2-21}
& Pt & Obj & Pt & Obj & Pt & Obj & Pt & Obj & Pt & Obj
& Pt & Obj & Pt & Obj & Pt & Obj & Pt & Obj & Pt & Obj \\
\midrule
BTF (Raw)
& 0.5206 & 0.6690
& 0.6044 & 0.5517
& 0.5173 & 0.7500
& 0.5186 & 0.4550
& 0.5833 & 0.5781
& 0.5523 & 0.4350
& 0.5399 & 0.4783
& 0.5534 & 0.4700
& 0.5986 & 0.4583
& 0.5613 & 0.6067 \\

BTF (FPFH)
& 0.5805 & 0.9850
& 0.6067 & 0.4917
& 0.6135 & 0.3183
& 0.5424 & 0.4400
& 0.6569 & 0.4609
& 0.7220 & 0.5583
& 0.6342 & 0.4867
& 0.5077 & 0.4717
& 0.5964 & 0.4700
& 0.4674 & 0.4967 \\

M3DM (Point-BERT)
& 0.6548 & \textbf{1.0000}
& 0.9180 & 0.6300
& 0.6159 & \textbf{0.9933}
& 0.6567 & 0.8983
& 0.9060 & 0.8203
& 0.8324 & 0.6767
& 0.8373 & \textbf{1.0000}
& 0.5465 & \textbf{0.9650}
& 0.7583 & 0.7083
& 0.6897 & \textbf{0.9483} \\

M3DM (Point-MAE)
& \textbf{0.6695} & \textbf{1.0000}
& \textbf{0.9442} & 0.5833
& 0.6624 & 0.9783
& 0.6999 & 0.9400
& 0.9396 & 0.6859
& \textbf{0.8538} & \textbf{0.6967}
& \textbf{0.8794} & \textbf{1.0000}
& 0.5927 & 0.9017
& 0.8101 & \textbf{0.8067}
& 0.7497 & 0.9283 \\

PatchCore (FPFH)
& 0.5072 & \textbf{1.0000}
& 0.7162 & 0.4650
& 0.5453 & 0.6183
& 0.4925 & 0.6150
& 0.2760 & 0.5047
& 0.4461 & 0.3350
& 0.5057 & 0.5333
& 0.4516 & 0.5700
& 0.4321 & 0.4533
& 0.4880 & 0.4450 \\

PatchCore (Point-MAE)
& 0.4491 & 0.6170
& 0.2433 & 0.4800
& 0.6017 & 0.7967
& 0.5633 & 0.5177
& 0.4328 & 0.4594
& 0.5317 & 0.4800
& 0.4833 & 0.5450
& 0.5083 & 0.4117
& 0.3900 & 0.5150
& 0.5017 & 0.6567 \\

Reg3D-AD
& 0.6281 & 0.8993
& 0.8177 & 0.6283
& 0.5871 & 0.8633
& 0.6097 & 0.8633
& 0.7785 & 0.6172
& 0.8008 & 0.6300
& 0.7950 & \textbf{1.0000}
& 0.5772 & 0.7283
& 0.6940 & 0.6817
& 0.6984 & 0.7683 \\

PO3AD
& 0.4794 & 0.5312
& 0.5685 & 0.4800
& 0.5071 & 0.2117
& 0.4825 & 0.4950
& 0.5159 & 0.3750
& 0.4920 & 0.4683
& 0.4881 & 0.5183
& 0.4838 & 0.4050
& 0.5107 & 0.4483
& 0.5353 & 0.3367 \\

PASDF
& 0.4980 & 0.6050
& 0.8285 & 0.4550
& 0.9219 & 0.4710
& 0.7108 & 0.5916
& 0.9279 & 0.6080
& 0.7378 & 0.6241
& 0.6074 & 0.5549
& 0.6876 & 0.5416
& 0.7241 & 0.4583
& 0.4861 & 0.6316 \\

Ours (SPA-SDF)
& 0.5949 & 0.9711
& 0.9159 & \textbf{0.7533}
& \textbf{0.9396} & 0.8600
& \textbf{0.8417} & \textbf{0.9833}
& \textbf{0.9707} & \textbf{0.8391}
& 0.6287 & 0.5583
& 0.8623 & 0.9050
& \textbf{0.6903} & 0.9167
& \textbf{0.9090} & 0.7800
& \textbf{0.7900} & 0.8550 \\
\midrule

\multirow{2}{*}{Method}
& \multicolumn{2}{c}{hinge3}
& \multicolumn{2}{c}{door3}
& \multicolumn{2}{c}{drawer3}
& \multicolumn{2}{c}{earphone\_case3}
& \multicolumn{2}{c}{fridge3}
& \multicolumn{2}{c}{injector3}
& \multicolumn{2}{c}{lamp3}
& \multicolumn{2}{c}{laptop3}
& \multicolumn{2}{c}{scissors3}
& \multicolumn{2}{c}{toilet\_seat3} \\
\cmidrule(lr){2-21}
& Pt & Obj & Pt & Obj & Pt & Obj & Pt & Obj & Pt & Obj
& Pt & Obj & Pt & Obj & Pt & Obj & Pt & Obj & Pt & Obj \\
\midrule
BTF (Raw)
& 0.5157 & 0.6053
& 0.4699 & 0.5817
& 0.4866 & 0.9750
& 0.5721 & 0.6450
& 0.5958 & 0.5217
& 0.5824 & 0.5450
& 0.5426 & 0.5967
& 0.5536 & 0.4917
& 0.5591 & 0.3617
& 0.5815 & 0.5750 \\

BTF (FPFH)
& 0.5483 & 0.5926
& 0.6521 & 0.6133
& 0.4969 & 0.9983
& 0.6193 & 0.3183
& 0.6452 & 0.5683
& 0.6306 & 0.5633
& 0.5102 & 0.5283
& 0.5256 & 0.6017
& 0.5321 & 0.5617
& 0.6110 & 0.5560 \\

M3DM (Point-BERT)
& 0.5961 & \textbf{1.0000}
& 0.8499 & 0.6833
& 0.5548 & \textbf{1.0000}
& 0.6847 & 0.8783
& 0.9177 & 0.5950
& 0.7709 & 0.6983
& 0.7848 & 0.9183
& 0.6200 & \textbf{0.9050}
& 0.7665 & 0.8417
& 0.7188 & \textbf{0.9167} \\

M3DM (Point-MAE)
& 0.6199 & \textbf{1.0000}
& 0.8941 & 0.6833
& 0.5513 & \textbf{1.0000}
& 0.7178 & 0.9250
& 0.9381 & 0.6100
& 0.7848 & 0.7383
& 0.8477 & \textbf{0.9667}
& 0.6463 & 0.8133
& 0.8028 & \textbf{0.8433}
& 0.7920 & 0.8983 \\

PatchCore (FPFH)
& 0.4636 & 0.8889
& 0.6701 & 0.5267
& 0.4272 & \textbf{1.0000}
& 0.5411 & 0.5633
& 0.6804 & 0.6967
& 0.4254 & 0.2700
& 0.4835 & 0.4833
& 0.6055 & 0.5683
& 0.5106 & 0.5933
& 0.4326 & 0.5350 \\

PatchCore (Point-MAE)
& 0.4625 & 0.5071
& 0.4133 & 0.5367
& 0.4783 & 0.5283
& 0.4033 & 0.5250
& 0.5250 & 0.5083
& 0.6150 & 0.4170
& 0.3833 & 0.6150
& 0.4133 & 0.4033
& 0.3967 & 0.5333
& 0.4267 & 0.6183 \\

Reg3D-AD
& 0.5597 & 0.8299
& 0.7231 & 0.6517
& 0.5163 & \textbf{1.0000}
& 0.6169 & 0.6600
& 0.8275 & 0.7100
& 0.7089 & 0.5800
& 0.7225 & 0.9067
& 0.5360 & 0.7033
& 0.6925 & 0.8017
& 0.6457 & 0.7217 \\

PO3AD
& 0.5378 & 0.1296
& 0.5100 & 0.4117
& 0.5090 & 0.4050
& 0.5228 & 0.4783
& 0.4447 & 0.5333
& 0.3909 & 0.4450
& 0.4409 & 0.4417
& 0.4880 & 0.5800
& 0.5953 & 0.5467
& 0.5327 & 0.4250 \\

PASDF
& 0.5030 & 0.5532
& 0.7963 & 0.5016
& 0.9678 & 0.5383
& 0.5480 & 0.5350
& 0.9580 & 0.5467
& \textbf{0.9647} & 0.5982
& 0.8210 & 0.6830
& 0.5780 & 0.5833
& 0.6006 & 0.5160
& 0.6923 & 0.5460 \\

Ours (SPA-SDF)
& \textbf{0.6255} & 0.9340
& \textbf{0.9123} & \textbf{0.7783}
& \textbf{0.9851} & 0.9850
& \textbf{0.7764} & \textbf{0.9650}
& \textbf{0.9891} & \textbf{0.8533}
& 0.7767 & \textbf{0.8450}
& \textbf{0.9382} & 0.9617
& \textbf{0.6933} & 0.8850
& \textbf{0.8912} & 0.8333
& \textbf{0.9364} & 0.9133 \\
\midrule

\multirow{2}{*}{Method}
& \multicolumn{2}{c}{usb}
& \multicolumn{2}{c}{usb2}
& \multicolumn{2}{c}{usb3}
& \multicolumn{2}{c}{utility\_knife}
& \multicolumn{2}{c}{utility\_knife2}
& \multicolumn{2}{c}{utility\_knife3}
& \multicolumn{2}{c}{folding\_knife}
& \multicolumn{2}{c}{folding\_knife2}
& \multicolumn{2}{c}{folding\_knife3}
& \multicolumn{2}{c}{Mean} \\
\cmidrule(lr){2-21}
& Pt & Obj & Pt & Obj & Pt & Obj & Pt & Obj & Pt & Obj
& Pt & Obj & Pt & Obj & Pt & Obj & Pt & Obj & Pt & Obj \\
\midrule
BTF (Raw)
& 0.5419 & 0.4439
& 0.4826 & 0.4933
& 0.4931 & 0.3933
& 0.5670 & 0.5484
& 0.5700 & 0.5250
& 0.5396 & 0.5312
& 0.6051 & 0.4541
& 0.5882 & 0.5133
& 0.5888 & 0.5350
& 0.5485 & 0.5522 \\

BTF (FPFH)
& 0.5227 & 0.3463
& 0.5383 & 0.5500
& 0.5198 & 0.6350
& 0.5629 & 0.9677
& 0.5175 & 0.8067
& 0.5176 & 0.5031
& 0.6396 & 0.4243
& 0.6066 & 0.5917
& 0.6441 & 0.4883
& 0.5746 & 0.5880 \\

M3DM (Point-BERT)
& 0.6926 & 0.9110
& 0.5328 & 0.7000
& 0.5780 & 0.6217
& 0.6508 & \textbf{1.0000}
& 0.5713 & \textbf{1.0000}
& 0.6905 & 0.7594
& 0.7759 & 0.9527
& 0.6741 & 0.9467
& 0.7056 & \textbf{1.0000}
& 0.7074 & 0.8655 \\

M3DM (Point-MAE)
& 0.7955 & 0.9360
& 0.5939 & 0.6750
& 0.6046 & 0.8017
& 0.6661 & \textbf{1.0000}
& 0.5981 & \textbf{1.0000}
& 0.6845 & 0.7906
& 0.8343 & \textbf{0.9986}
& 0.7516 & \textbf{1.0000}
& 0.7534 & \textbf{1.0000}
& 0.7468 & 0.8714 \\

PatchCore (FPFH)
& 0.4507 & 0.6524
& 0.4447 & 0.6583
& 0.4988 & 0.5767
& 0.5211 & 0.9984
& 0.4901 & \textbf{1.0000}
& 0.5084 & 0.6125
& 0.5619 & 0.5760
& 0.5757 & 0.6483
& 0.4790 & 0.6817
& 0.5090 & 0.6424 \\

PatchCore (Point-MAE)
& 0.4507 & 0.5720
& 0.4470 & 0.4700
& 0.3983 & 0.5333
& 0.6177 & \textbf{1.0000}
& 0.5267 & 0.8717
& 0.5084 & 0.6125
& 0.5243 & 0.4824
& 0.5383 & 0.3767
& 0.4790 & 0.6170
& 0.4828 & 0.5636 \\

Reg3D-AD
& 0.5953 & 0.6829
& 0.5116 & 0.5533
& 0.5197 & 0.4467
& 0.6467 & \textbf{1.0000}
& 0.5821 & \textbf{1.0000}
& 0.5976 & 0.6641
& 0.6962 & 0.7446
& 0.6031 & 0.8867
& 0.6106 & 0.7067
& 0.6531 & 0.7663 \\

PO3AD
& 0.5297 & 0.5683
& 0.4943 & 0.4367
& 0.5319 & 0.5550
& 0.5071 & 0.3516
& 0.5248 & 0.4367
& 0.5431 & 0.4453
& 0.3958 & 0.4189
& 0.4693 & 0.3167
& 0.4690 & 0.4450
& 0.4979 & 0.4672 \\

PASDF
& 0.7650 & 0.5817
& 0.6226 & 0.5560
& 0.7357 & 0.5416
& \textbf{0.7812} & 0.8467
& 0.6493 & 0.6916
& 0.7910 & 0.5249
& 0.7337 & 0.5700
& 0.7103 & 0.6760
& 0.7337 & 0.5700
& 0.6950 & 0.6890 \\

Ours (SPA-SDF)
& \textbf{0.9859} & \textbf{0.9829}
& \textbf{0.8364} & \textbf{0.8017}
& \textbf{0.8061} & \textbf{0.8817}
& 0.7680 & 0.7823
& \textbf{0.7165} & 0.6017
& \textbf{0.8373} & \textbf{0.9297}
& \textbf{0.9179} & 0.8541
& \textbf{0.8741} & 0.9267
& \textbf{0.9274} & 0.8900
& \textbf{0.8410} & \textbf{0.8740} \\
\bottomrule
\end{tabular}
\end{adjustbox}
\end{table*}

\section{Ablation Study}

\subsection{Component Ablation Study}

We first analyze the contribution of each component in the proposed SPA-SDF framework.

Table~\ref{tab:ablation1_seen} reports the results in the Seen setting.
Table~\ref{tab:ablation1_unseen} presents the corresponding results in the Unseen setting.

The results show that each component consistently improves performance,
and removing any module leads to noticeable degradation.
In particular, the pose-conditioning module plays a critical role in handling articulated variations.

\subsection{Pose Modeling Analysis}

We further investigate the effect of pose modeling on anomaly detection.

Table~\ref{tab:ablation2_seen} summarizes the results in the Seen setting,
while Table~\ref{tab:ablation2_unseen} reports the results in the Unseen setting.

Compared to variants without explicit pose modeling, our full model achieves
significantly better performance, especially under the Unseen setting,
highlighting the importance of modeling pose-dependent geometry.

\section{Additional Qualitative Results}

\begin{figure*}[!t]
\centering

\begin{subfigure}{\textwidth}
    \centering
    \includegraphics[width=\textwidth]{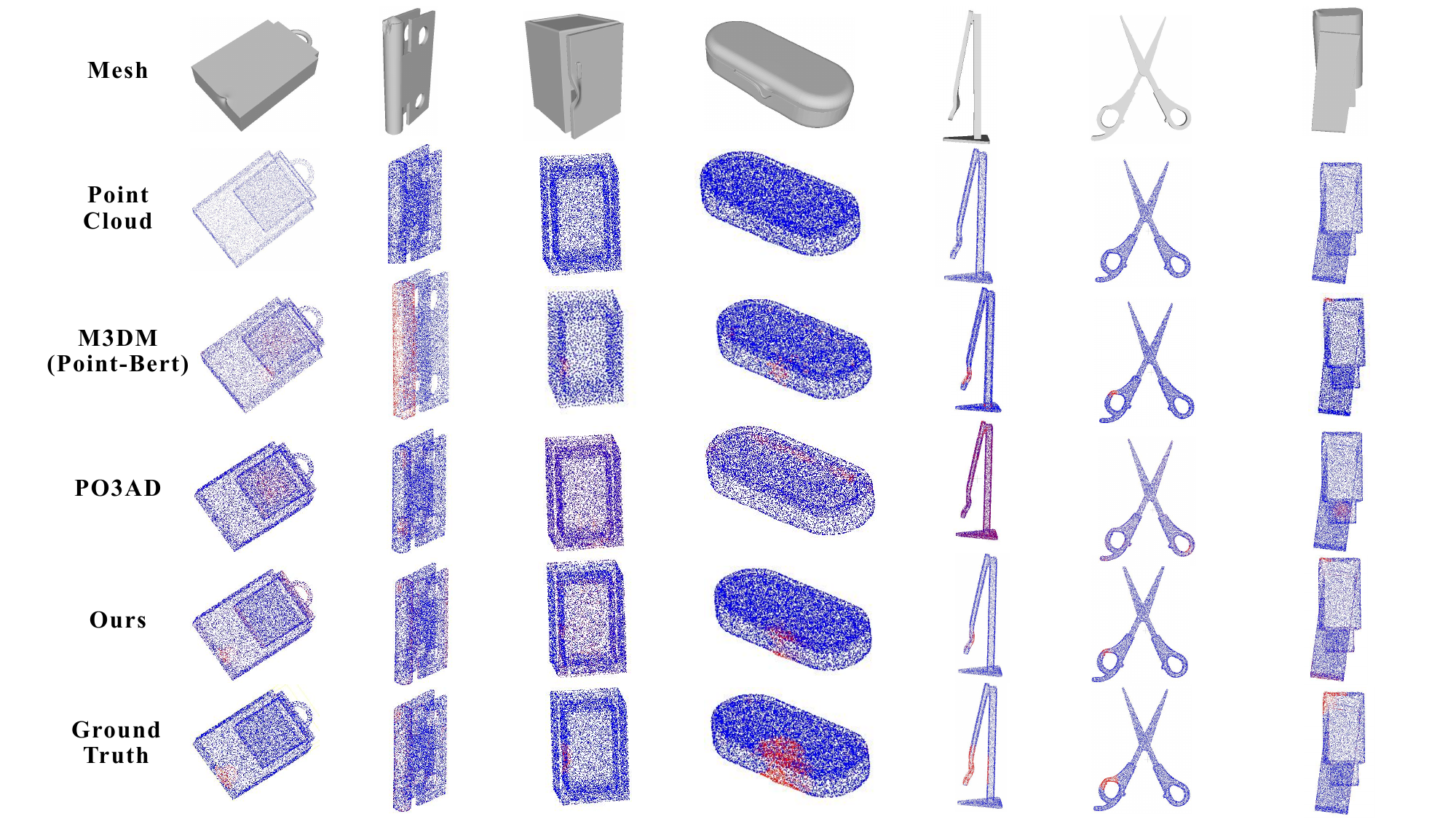}
    \caption{Qualitative comparison with existing methods under articulated variations, where SPA-SDF achieves more accurate and coherent anomaly localization than M3DM and PO3AD.}
    \label{fig:supp_compare}
\end{subfigure}

\vspace{2mm}

\begin{subfigure}{\textwidth}
    \centering
    \includegraphics[width=\textwidth]{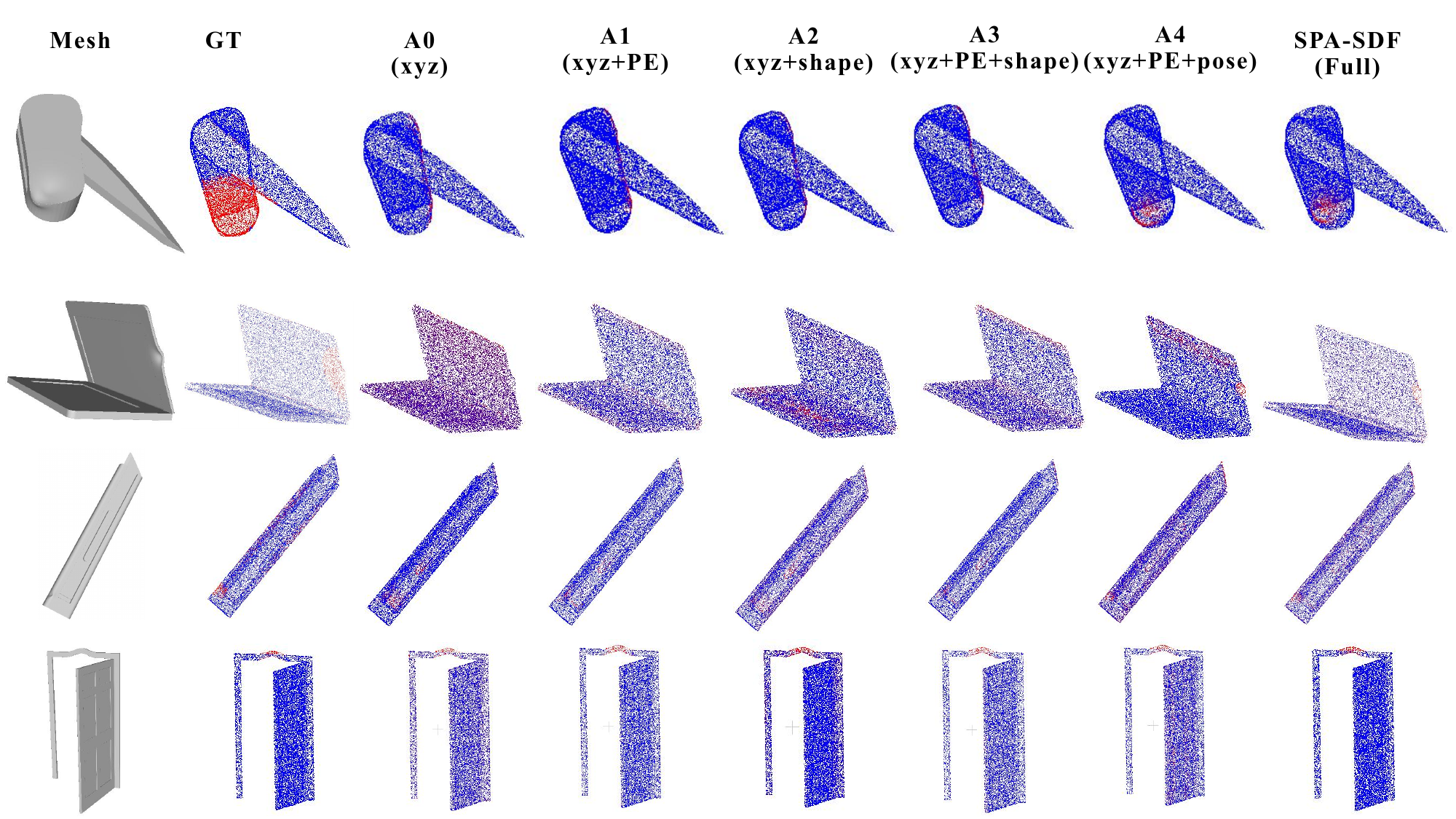}
    \caption{Qualitative ablation of articulation modeling, showing that breaking the rigid prior progressively improves anomaly localization under articulated states.}
    \label{fig:supp_pose}
\end{subfigure}

\vspace{-2mm}
\caption{Additional qualitative results, including comparisons with existing methods and articulation-focused ablations.}
\label{fig:supp_qualitative}
\vspace{-2mm}
\end{figure*}

Beyond the quantitative benchmarks reported above, we provide additional qualitative evidence to further dissect the behavior of SPA-SDF.
Through visual comparisons (Fig.~\ref{fig:supp_compare}) and articulation-focused ablations (Fig.~\ref{fig:supp_pose}), we demonstrate the model's precision under diverse articulated states.
This fine-grained perspective confirms that bypassing the rigid prior through explicit articulation modeling is essential for reliable anomaly localization in articulated scenarios.

\clearpage
\begin{table*}[t]
\centering
\caption{Component ablation study on \textbf{seen} classes. Best per column in bold. }
\label{tab:ablation1_seen}
\footnotesize
\setlength{\tabcolsep}{1.5pt}
\begin{adjustbox}{width=\textwidth}
\begin{tabular}{l*{10}{cc}}
\toprule
\toprule
\multirow{2}{*}{Method}
& \multicolumn{2}{c}{hinge}
& \multicolumn{2}{c}{door}
& \multicolumn{2}{c}{drawer}
& \multicolumn{2}{c}{earphone\_case}
& \multicolumn{2}{c}{fridge}
& \multicolumn{2}{c}{injector}
& \multicolumn{2}{c}{lamp}
& \multicolumn{2}{c}{laptop}
& \multicolumn{2}{c}{scissors}
& \multicolumn{2}{c}{toilet\_seat} \\
\cmidrule(lr){2-21}
& Pt & Obj & Pt & Obj & Pt & Obj & Pt & Obj & Pt & Obj
& Pt & Obj & Pt & Obj & Pt & Obj & Pt & Obj & Pt & Obj \\
\midrule

A0 (xyz)
& 0.5798 & 0.628
& 0.6123 & 0.5377
& 0.8246 & 0.1543
& 0.7607 & 0.9166
& 0.8070 & 0.8721
& 0.8640 & 0.9233
& 0.6769 & 0.8183
& 0.6887 & 0.8533
& 0.6566 & 0.6308
& 0.6640 & 0.5461 \\

A1 (xyz+PE)
& 0.5977 & 0.613
& 0.6219 & 0.7114
& 0.9441 & 0.361
& 0.8274 & 0.9299
& 0.8761 & 0.8605
& 0.9860 & 0.9666
& 0.8484 & 0.899
& 0.6586 & 0.6283
& 0.7720 & 0.649
& 0.5037 & 0.7046 \\

A2 (xyz+shape)
& 0.5630 & 0.645
& 0.6174 & 0.537
& 0.7711 & 0.278
& 0.7584 & 0.933
& 0.8590 & 0.8721
& 0.8305 & 0.926
& 0.7190 & 0.8283
& 0.6513 & 0.8316
& 0.6219 & 0.735
& 0.5595 & 0.6484 \\

A3 (xyz+PE+shape)
& 0.5807 & 0.638
& 0.6722 & 0.6524
& \textbf{0.9485} & 0.421
& 0.8401 & 0.9233
& 0.8852 & 0.9163
& 0.9883 & 0.9783
& 0.7037 & 0.7833
& 0.6412 & 0.6366
& 0.6324 & 0.6583
& 0.5377 & 0.7093 \\

A4 (xyz+PE+pose)
& 0.8486 & \textbf{0.967}
& 0.8294 & 0.7606
& 0.9042 & 0.854
& \textbf{0.9240} & 0.9874
& 0.9836 & 0.969
& 0.2170 & 0.978
& 0.7982 & 0.889
& \textbf{0.9933} & \textbf{1.000}
& \textbf{0.9149} & 0.768
& \textbf{0.9328} & \textbf{0.988} \\

Full SPA-SDF
& \textbf{0.9472} & 0.9646
& \textbf{0.8963} & \textbf{0.9623}
& 0.9228 & \textbf{0.8655}
& 0.9219 & \textbf{0.995}
& \textbf{0.993} & \textbf{1.000}
& \textbf{0.9963} & \textbf{1.000}
& \textbf{0.9395} & \textbf{0.983}
& 0.8792 & \textbf{1.000}
& 0.9030 & \textbf{0.933}
& 0.7359 & 0.9469 \\

\midrule
\toprule
\multirow{2}{*}{Method}
& \multicolumn{2}{c}{hinge2}
& \multicolumn{2}{c}{door2}
& \multicolumn{2}{c}{drawer2}
& \multicolumn{2}{c}{earphone\_case2}
& \multicolumn{2}{c}{fridge2}
& \multicolumn{2}{c}{injector2}
& \multicolumn{2}{c}{lamp2}
& \multicolumn{2}{c}{laptop2}
& \multicolumn{2}{c}{scissors2}
& \multicolumn{2}{c}{toilet\_seat2} \\
\cmidrule(lr){2-21}
& Pt & Obj & Pt & Obj & Pt & Obj & Pt & Obj & Pt & Obj
& Pt & Obj & Pt & Obj & Pt & Obj & Pt & Obj & Pt & Obj \\
\midrule

A0 (xyz)
& 0.4794 & 0.553
& 0.8114 & 0.6616
& 0.7140 & 0.740
& 0.5676 & 0.6966
& 0.8055 & 0.768
& 0.6198 & 0.6642
& 0.6102 & 0.6516
& 0.5980 & 0.5283
& 0.5450 & 0.580
& 0.6206 & 0.625 \\

A1 (xyz+PE)
& 0.4980 & 0.5416
& 0.8346 & 0.575
& \textbf{0.9494} & 0.606
& 0.7059 & 0.6583
& 0.9193 & 0.6263
& 0.6228 & 0.6363
& 0.6702 & 0.6666
& 0.6794 & 0.516
& 0.7484 & 0.470
& 0.5767 & 0.5666 \\

A2 (xyz+shape)
& 0.4876 & 0.5425
& 0.8226 & 0.6083
& 0.7233 & 0.679
& 0.5421 & 0.7583
& 0.8493 & 0.6919
& 0.6049 & \textbf{0.6699}
& 0.5910 & 0.568
& 0.6086 & 0.630
& 0.6569 & 0.583
& 0.5933 & 0.642 \\

A3 (xyz+PE+shape)
& 0.5720 & 0.5625
& 0.8723 & 0.6616
& 0.9393 & 0.418
& 0.6370 & 0.6666
& 0.8035 & 0.6194
& 0.6268 & 0.6265
& 0.6630 & 0.613
& 0.6450 & 0.560
& 0.6953 & 0.5283
& 0.5860 & 0.616 \\

A4 (xyz+PE+pose)
& 0.5923 & \textbf{0.976}
& 0.9192 & 0.7516
& 0.9461 & \textbf{0.9083}
& \textbf{0.8291} & 0.970
& 0.9847 & \textbf{0.920}
& 0.7535 & 0.405
& 0.8418 & \textbf{0.941}
& 0.7480 & 0.816
& \textbf{0.9080} & \textbf{0.733}
& 0.7283 & 0.8366 \\

Full SPA-SDF
& \textbf{0.6012} & 0.9479
& \textbf{0.982} & \textbf{0.952}
& 0.9427 & 0.905
& 0.807 & \textbf{0.9717}
& \textbf{0.9862} & 0.754
& \textbf{0.773} & 0.4701
& \textbf{0.8692} & 0.9267
& \textbf{0.7622} & \textbf{0.8933}
& 0.9065 & \textbf{0.733}
& \textbf{0.762} & \textbf{0.883} \\

\midrule
\toprule
\multirow{2}{*}{Method}
& \multicolumn{2}{c}{hinge3}
& \multicolumn{2}{c}{door3}
& \multicolumn{2}{c}{drawer3}
& \multicolumn{2}{c}{earphone\_case3}
& \multicolumn{2}{c}{fridge3}
& \multicolumn{2}{c}{injector3}
& \multicolumn{2}{c}{lamp3}
& \multicolumn{2}{c}{laptop3}
& \multicolumn{2}{c}{scissors3}
& \multicolumn{2}{c}{toilet\_seat3} \\
\cmidrule(lr){2-21}
& Pt & Obj & Pt & Obj & Pt & Obj & Pt & Obj & Pt & Obj
& Pt & Obj & Pt & Obj & Pt & Obj & Pt & Obj & Pt & Obj \\
\midrule

A0 (xyz)
& 0.4995 & 0.611
& 0.8841 & 0.6166
& 0.7873 & 0.3916
& 0.5440 & 0.596
& 0.8082 & 0.7683
& 0.9250 & \textbf{0.6086}
& 0.7048 & 0.703
& 0.5147 & 0.6649
& 0.6236 & 0.5316
& 0.1880 & 0.6983 \\

A1 (xyz+PE)
& 0.5420 & 0.5717
& 0.8899 & 0.5133
& 0.9780 & 0.5316
& 0.5624 & 0.510
& 0.9525 & 0.6477
& 0.9625 & 0.6059
& 0.8178 & 0.7533
& 0.5405 & 0.570
& 0.6055 & 0.500
& 0.6580 & 0.6649 \\

A2 (xyz+shape)
& 0.4782 & 0.510
& 0.8669 & 0.6633
& 0.8924 & 0.350
& 0.5283 & 0.551
& 0.8770 & 0.739
& 0.9320 & 0.6049
& 0.7784 & 0.696
& 0.5300 & 0.730
& 0.5842 & 0.550
& 0.6870 & 0.765 \\

A3 (xyz+PE+shape)
& 0.5395 & 0.5717
& 0.9087 & 0.5383
& \textbf{0.9821} & 0.549
& 0.5417 & 0.516
& 0.8654 & 0.689
& \textbf{0.9637} & 0.604
& 0.7518 & 0.766
& 0.5789 & 0.585
& 0.6767 & 0.5283
& 0.7960 & 0.6233 \\

A4 (xyz+PE+pose)
& \textbf{0.7195} & \textbf{0.974}
& \textbf{0.9540} & 0.6816
& 0.9729 & \textbf{0.958}
& \textbf{0.7041} & \textbf{0.946}
& 0.9876 & \textbf{0.908}
& 0.9316 & 0.5846
& \textbf{0.9284} & 0.916
& \textbf{0.7160} & 0.833
& 0.8454 & 0.726
& 0.8981 & 0.950 \\

Full SPA-SDF
& 0.696 & 0.944
& 0.865 & \textbf{0.9992}
& 0.8935 & 0.6083
& 0.6678 & 0.917
& \textbf{0.9955} & 0.877
& 0.9207 & 0.444
& 0.9018 & \textbf{0.956}
& 0.6929 & \textbf{0.8833}
& \textbf{0.863} & \textbf{0.765}
& \textbf{0.925} & \textbf{0.9567} \\

\midrule
\toprule
\multirow{2}{*}{Method}
& \multicolumn{2}{c}{usb}
& \multicolumn{2}{c}{usb2}
& \multicolumn{2}{c}{usb3}
& \multicolumn{2}{c}{utility\_knife}
& \multicolumn{2}{c}{utility\_knife2}
& \multicolumn{2}{c}{utility\_knife3}
& \multicolumn{2}{c}{folding\_knife}
& \multicolumn{2}{c}{folding\_knife2}
& \multicolumn{2}{c}{folding\_knife3}
& \multicolumn{2}{c}{Mean} \\
\cmidrule(lr){2-21}
& Pt & Obj & Pt & Obj & Pt & Obj & Pt & Obj & Pt & Obj
& Pt & Obj & Pt & Obj & Pt & Obj & Pt & Obj & Pt & Obj \\
\midrule

A0 (xyz)
& 0.5091&0.4708
& 0.5390&0.593
& 0.5810&0.468
& 0.6687&0.6501
& 0.5506&0.744
& 0.6585&0.4562
& 0.6455&0.619
& 0.4936&0.585
& 0.6455&0.619
& 0.5981 & 0.6335 \\

A1 (xyz+PE)
& 0.5110&0.5799
& 0.6540&0.5716
& 0.7509&0.440
& 0.8050&0.667
& 0.6724&0.715
& 0.7849&0.4531
& 0.6838&0.616
& 0.6964&0.5716
& 0.6838&0.616
& 0.6621 & 0.6847 \\

A2 (xyz+shape)
& 0.5091&0.4708
& 0.5524&0.6066
& 0.5709&0.4283
& 0.7035&0.6895
& 0.6480&0.7457
& 0.6397&0.4625
& 0.6315&0.586
& 0.5524&0.4983
& 0.6315&0.586
& 0.6393 & 0.6759 \\

A3 (xyz+PE+shape)
& 0.5110&0.5799
& 0.7290&0.5533
& 0.7962&0.483
& 0.8329&0.6989
& 0.7150&0.722
& 0.8265&0.4437
& 0.8097&0.626
& 0.7135&0.6166
& 0.8097&0.626
& 0.6889 & 0.7070 \\

A4 (xyz+PE+pose)
& \textbf{0.7923}&\textbf{0.9630}
& 0.6997&0.5816
& 0.6482&0.5883
& \textbf{0.9880}&\textbf{0.9927}
& 0.6632&0.727
& 0.8117&0.8484
& 0.8413&0.918
& 0.8389&0.785
& 0.8413&0.918
& 0.8302 & 0.8520 \\

Full SPA-SDF
& \textbf{0.7923}&\textbf{0.9630}
& \textbf{0.8276}&\textbf{0.7617}
& \textbf{0.8988}&\textbf{0.7967}
& 0.9728&0.9781
& \textbf{0.7379}&\textbf{0.8831}
& \textbf{0.8390}&\textbf{0.8578}
& \textbf{0.915}&\textbf{0.956}
& \textbf{0.8947}&\textbf{0.900}
& \textbf{0.915}&\textbf{0.956}
& \textbf{0.8651} & \textbf{0.8843} \\

\bottomrule
\end{tabular}
\end{adjustbox}
\end{table*}

\begin{table*}[t]
\centering
\caption{Component ablation study on \textbf{unseen} classes. Best per column in bold.}
\label{tab:ablation1_unseen}
\footnotesize
\setlength{\tabcolsep}{2.2pt}
\begin{adjustbox}{width=\textwidth}
\begin{tabular}{l*{10}{cc}}
\toprule
\toprule
\multirow{2}{*}{Method}
& \multicolumn{2}{c}{hinge}
& \multicolumn{2}{c}{door}
& \multicolumn{2}{c}{drawer}
& \multicolumn{2}{c}{earphone\_case}
& \multicolumn{2}{c}{fridge}
& \multicolumn{2}{c}{injector}
& \multicolumn{2}{c}{lamp}
& \multicolumn{2}{c}{laptop}
& \multicolumn{2}{c}{scissors}
& \multicolumn{2}{c}{toilet\_seat} \\
\cmidrule(lr){2-21}
& Pt & Obj & Pt & Obj & Pt & Obj & Pt & Obj & Pt & Obj
& Pt & Obj & Pt & Obj & Pt & Obj & Pt & Obj & Pt & Obj \\
\midrule

A0 (xyz)
& 0.5639 & 0.626
& 0.4873 & 0.5296
& 0.8128 & 0.173
& 0.7253 & 0.930
& 0.8614 & 0.8707
& 0.7629 & 0.8694
& 0.6789 & 0.783
& 0.6921 & 0.7048
& 0.6403 & 0.715
& 0.4967 & 0.6255 \\

A1 (xyz+PE)
& 0.5660 & 0.585
& 0.6530 & 0.633
& \textbf{0.9450} & 0.395
& 0.7950 & 0.909
& 0.8630 & 0.629
& 0.9050 & 0.953
& 0.8530 & 0.816
& 0.7040 & 0.6258
& 0.6776 & 0.695
& 0.5310 & 0.6647 \\

A2 (xyz+shape)
& 0.5165 & 0.700
& 0.5132 & 0.5018
& 0.8013 & 0.285
& 0.7070 & 0.9316
& 0.8520 & 0.763
& 0.7881 & 0.8813
& 0.6775 & 0.749
& 0.6571 & 0.7048
& 0.6463 & 0.723
& 0.4857 & 0.6044 \\

A3 (xyz+PE+shape)
& 0.5683 & 0.595
& 0.6676 & 0.6703
& 0.9442 & 0.441
& 0.8088 & 0.929
& 0.8786 & 0.694
& 0.9535 & 0.971
& 0.7120 & 0.813
& 0.6903 & 0.5967
& 0.6746 & 0.693
& 0.5480 & 0.672 \\

A4 (xyz+PE+pose)
& 0.8571 & 0.8603
& 0.6879 & 0.6222
& 0.8790 & 0.6043
& \textbf{0.9025} & \textbf{0.990}
& 0.9726 & 0.9715
& \textbf{0.9729} & \textbf{0.9796}
& \textbf{0.9029} & 0.906
& \textbf{0.7676} & 0.612
& 0.8536 & 0.770
& 0.6176 & 0.7176 \\

Full SPA-SDF
& \textbf{0.9599} & \textbf{0.9388}
& \textbf{0.8489} & \textbf{0.9444}
& 0.9396 & \textbf{0.8600}
& 0.8749 & 0.9317
& \textbf{0.9911} & \textbf{0.9887}
& 0.6000 & 0.6304
& 0.8959 & \textbf{0.9800}
& 0.7535 & \textbf{0.9226}
& \textbf{0.8844} & \textbf{0.9600}
& \textbf{0.7091} & \textbf{0.9132} \\

\midrule
\toprule
\multirow{2}{*}{Method}
& \multicolumn{2}{c}{hinge2}
& \multicolumn{2}{c}{door2}
& \multicolumn{2}{c}{drawer2}
& \multicolumn{2}{c}{earphone\_case2}
& \multicolumn{2}{c}{fridge2}
& \multicolumn{2}{c}{injector2}
& \multicolumn{2}{c}{lamp2}
& \multicolumn{2}{c}{laptop2}
& \multicolumn{2}{c}{scissors2}
& \multicolumn{2}{c}{toilet\_seat2} \\
\cmidrule(lr){2-21}
& Pt & Obj & Pt & Obj & Pt & Obj & Pt & Obj & Pt & Obj
& Pt & Obj & Pt & Obj & Pt & Obj & Pt & Obj & Pt & Obj \\
\midrule

A0 (xyz)
& 0.4736 & 0.567
& 0.839 & 0.69
& 0.781 & 0.683
& 0.569 & 0.535
& 0.848 & 0.8297
& 0.7348 & 0.6225
& 0.519 & 0.616
& 0.69 & 0.6166
& 0.6066 & 0.5083
& 0.6585 & 0.5933 \\

A1 (xyz+PE)
& 0.498 & 0.605
& 0.8285 & 0.455
& 0.9219 & 0.471
& 0.7108 & 0.5916
& 0.9279 & 0.608
& 0.7378 & 0.6241
& 0.6074 & 0.5549
& 0.6876 & 0.5416
& 0.7241 & 0.4583
& 0.4861 & 0.6316 \\

A2 (xyz+shape)
& 0.4725 & 0.5648
& 0.8408 & 0.588
& 0.749 & 0.645
& 0.5736 & 0.656
& 0.8736 & 0.7058
& 0.7343 & \textbf{0.6356}
& 0.6039 & 0.5416
& 0.628 & 0.6183
& 0.729 & 0.4883
& 0.6 & 0.6166 \\

A3 (xyz+PE+shape)
& 0.569 & 0.5648
& 0.866 & 0.439
& 0.9276 & 0.4
& 0.5183 & 0.6574
& 0.832 & 0.6131
& 0.7556 & 0.6192
& 0.612 & 0.5566
& 0.6636 & 0.536
& 0.7149 & 0.4466
& 0.437 & 0.6433 \\

A4 (xyz+PE+pose)
& 0.5355 & 0.877
& 0.9151 & 0.6783
& 0.9364 & 0.706
& 0.795 & 0.863
& \textbf{0.9707} & 0.715
& \textbf{0.7972} & 0.5204
& 0.6648 & 0.791
& 0.6753 & 0.7583
& 0.8643 & 0.691
& 0.7254 & 0.801 \\

Full SPA-SDF
& \textbf{0.5949} & \textbf{0.9711}
& \textbf{0.9159} & \textbf{0.7533}
& \textbf{0.9396} & \textbf{0.8600}
& \textbf{0.8417} & \textbf{0.9833}
& \textbf{0.9707} & \textbf{0.8391}
& 0.6287 & 0.5583
& \textbf{0.8623} & \textbf{0.9050}
& \textbf{0.6903} & \textbf{0.9167}
& \textbf{0.9090} & \textbf{0.7800}
& \textbf{0.7900} & \textbf{0.8550} \\

\midrule
\toprule
\multirow{2}{*}{Method}
& \multicolumn{2}{c}{hinge3}
& \multicolumn{2}{c}{door3}
& \multicolumn{2}{c}{drawer3}
& \multicolumn{2}{c}{earphone\_case3}
& \multicolumn{2}{c}{fridge3}
& \multicolumn{2}{c}{injector3}
& \multicolumn{2}{c}{lamp3}
& \multicolumn{2}{c}{laptop3}
& \multicolumn{2}{c}{scissors3}
& \multicolumn{2}{c}{toilet\_seat3} \\
\cmidrule(lr){2-21}
& Pt & Obj & Pt & Obj & Pt & Obj & Pt & Obj & Pt & Obj
& Pt & Obj & Pt & Obj & Pt & Obj & Pt & Obj & Pt & Obj \\
\midrule

A0 (xyz)
& 0.5349 & 0.5995
& 0.798 & 0.700
& 0.856 & 0.4183
& 0.5967 & 0.665
& 0.857 & 0.777
& 0.9357 & 0.6058
& 0.7113 & 0.7183
& 0.5417 & 0.5483
& 0.6167 & 0.4967
& 0.822 & 0.6666 \\

A1 (xyz+PE)
& 0.503 & 0.5532
& 0.7963 & 0.5016
& 0.9678 & 0.5383
& 0.548 & 0.535
& 0.958 & 0.5467
& 0.9647 & 0.5982
& 0.821 & 0.683
& 0.578 & 0.5833
& 0.6006 & 0.516
& 0.6923 & 0.546 \\

A2 (xyz+shape)
& 0.5321 & 0.5185
& 0.774 & 0.555
& 0.8505 & 0.310
& 0.6054 & 0.613
& 0.8965 & 0.543
& 0.9393 & 0.6084
& 0.7602 & 0.728
& 0.5322 & 0.5483
& 0.6051 & 0.572
& 0.7198 & 0.7066 \\

A3 (xyz+PE+shape)
& 0.5002 & 0.560
& 0.852 & 0.583
& 0.968 & 0.5416
& 0.538 & 0.6366
& 0.8965 & 0.5439
& \textbf{0.9652} & 0.5829
& 0.7647 & 0.656
& 0.559 & 0.5083
& 0.6635 & 0.440
& 0.7983 & 0.5533 \\

A4 (xyz+PE+pose)
& \textbf{0.6305} & 0.8946
& 0.8754 & 0.7016
& 0.9824 & 0.940
& 0.736 & 0.8133
& 0.9806 & 0.790
& 0.9161 & 0.4872
& 0.899 & 0.833
& 0.6546 & 0.8466
& 0.8544 & 0.700
& 0.8561 & 0.8949 \\

Full SPA-SDF
& 0.6255 & \textbf{0.9340}
& \textbf{0.9123} & \textbf{0.7783}
& \textbf{0.9851} & \textbf{0.9850}
& \textbf{0.7764} & \textbf{0.9650}
& \textbf{0.9891} & \textbf{0.8533}
& 0.7767 & \textbf{0.8450}
& \textbf{0.9382} & \textbf{0.9617}
& \textbf{0.6933} & \textbf{0.8850}
& \textbf{0.8912} & \textbf{0.8333}
& \textbf{0.9364} & \textbf{0.9133} \\

\midrule
\toprule
\multirow{2}{*}{Method}
& \multicolumn{2}{c}{usb}
& \multicolumn{2}{c}{usb2}
& \multicolumn{2}{c}{usb3}
& \multicolumn{2}{c}{utility\_knife}
& \multicolumn{2}{c}{utility\_knife2}
& \multicolumn{2}{c}{utility\_knife3}
& \multicolumn{2}{c}{folding\_knife}
& \multicolumn{2}{c}{folding\_knife2}
& \multicolumn{2}{c}{folding\_knife3}
& \multicolumn{2}{c}{Mean} \\
\cmidrule(lr){2-21}
& Pt & Obj & Pt & Obj & Pt & Obj & Pt & Obj & Pt & Obj
& Pt & Obj & Pt & Obj & Pt & Obj & Pt & Obj & Pt & Obj \\
\midrule

A0 (xyz)
& 0.671&0.597
& 0.529&0.578
& 0.5456&0.470
& 0.6398&0.8806
& 0.531&0.705
& 0.6567&0.489
& 0.7227&0.550
& 0.567&0.585
& 0.7227&0.550
& 0.6392 & 0.6664 \\

A1 (xyz+PE)
& 0.765&0.5817
& 0.6226&0.556
& 0.7357&0.5416
& 0.7812&0.8467
& 0.6493&0.6916
& 0.791&0.5249
& 0.7337&0.570
& 0.7103&0.676
& 0.7337&0.570
& 0.6952 &0.6891 \\

A2 (xyz+shape)
& 0.678&0.6365
& 0.546&0.595
& 0.4958&0.471
& 0.5618&0.9032
& 0.6507&\textbf{0.725}
& 0.6483&0.4968
& 0.6376&0.561
& 0.573&0.6499
& 0.6376&0.561
& 0.6663 & 0.6930 \\

A3 (xyz+PE+shape)
& 0.813&0.596
& 0.600&0.6947
& 0.781&0.550
& \textbf{0.8196}&0.8693
& 0.6973&0.6983
& 0.8362&0.5031
& 0.8218&0.563
& 0.7201&0.6083
& 0.8218&0.563
& 0.7140 & 0.7150 \\

A4 (xyz+PE+pose)
& 0.9517&0.954
& 0.659&0.5269
& 0.6877&0.5449
& 0.7807&\textbf{0.9806}
& 0.6383&0.6783
& 0.8155&0.7187
& 0.8766&0.763
& 0.8195&0.783
& 0.8766&0.763
& 0.8151 & 0.7701 \\

Full SPA-SDF
& \textbf{0.9859} & \textbf{0.9829}
& \textbf{0.8364} & \textbf{0.8017}
& \textbf{0.8061} & \textbf{0.8817}
& 0.7680 & 0.7823
& \textbf{0.7165} & 0.6017
& \textbf{0.8373} & \textbf{0.9297}
& \textbf{0.9179} & \textbf{0.8541}
& \textbf{0.8741} & \textbf{0.9267}
& \textbf{0.9274} & \textbf{0.8900}
& \textbf{0.8410} & \textbf{0.8740} \\

\bottomrule
\end{tabular}
\end{adjustbox}
\end{table*}

\begin{table*}[t]
\centering
\caption{Pose ablation on \textbf{seen} classes. Best per column in bold.}

\label{tab:ablation2_seen}
\footnotesize
\setlength{\tabcolsep}{2.2pt}
\begin{adjustbox}{width=\textwidth}
\begin{tabular}{l*{10}{cc}}
\toprule
\toprule
\multirow{2}{*}{Method}
& \multicolumn{2}{c}{hinge}
& \multicolumn{2}{c}{door}
& \multicolumn{2}{c}{drawer}
& \multicolumn{2}{c}{earphone\_case}
& \multicolumn{2}{c}{fridge}
& \multicolumn{2}{c}{injector}
& \multicolumn{2}{c}{lamp}
& \multicolumn{2}{c}{laptop}
& \multicolumn{2}{c}{scissors}
& \multicolumn{2}{c}{toilet\_seat} \\
\cmidrule(lr){2-21}
& Pt & Obj & Pt & Obj & Pt & Obj & Pt & Obj & Pt & Obj
& Pt & Obj & Pt & Obj & Pt & Obj & Pt & Obj & Pt & Obj \\
\midrule

A0 (w/o $\theta$)
& 0.5807 & 0.638
& 0.6722 & 0.6524
& \textbf{0.9485} & 0.421
& 0.8401 & 0.9233
& 0.8852 & 0.9163
& 0.9883 & 0.9783
& 0.7037 & 0.7833
& 0.6412 & 0.6366
& 0.6324 & 0.6583
& 0.5377 & 0.7093 \\

A1 (Raw $\theta$)
& 0.7732 & 0.8535
& 0.8238 & 0.8541
& 0.9178 & 0.7949
& 0.9082 & \textbf{1.0000}
& 0.9445 & 0.9967
& 0.6245 & 0.4339    
& 0.8794 & 0.9517
& 0.8884 & 0.9967
& 0.8646 & 0.8683
& 0.7201 & 0.9422 \\

A2 (Fixed $\sin/\cos$)
& 0.9084 & \textbf{0.9722}
& 0.8173 & 0.8656
& 0.9192 & 0.8668
& \textbf{0.9241} & 0.9900
& 0.9642 & 0.9917
& 0.6121 & 0.5089   
& 0.8791 & 0.9317
& \textbf{0.8977} & 0.99
& 0.8823 & 0.8933
& \textbf{0.7369} & \textbf{0.9469} \\

A3 (Fourier $L{=}8$)
& 0.9416 & 0.9672
& 0.8406 & 0.9000
& 0.9254 & 0.8531
& 0.8936 & 0.9900
& 0.9578 & 0.9820
& 0.6021 & 0.5446   
& 0.9117 & 0.9817
& 0.8489 & 0.9850
& 0.8937 & 0.9099
& 0.7162 & 0.9187 \\

A4 (Fourier $L{=}16$)
& 0.9446 & 0.9583
& 0.8149 & 0.8426
& 0.9235 & \textbf{0.8848}
& 0.9199 & 0.9900
& 0.9637 & 0.9778
& 0.6027 & 0.4125   
& 0.8897 & 0.9800
& 0.8663 & 0.98
& 0.8797 & 0.9133
& 0.7236 & 0.9438 \\

SPA-SDF (full)
& \textbf{0.9472} & 0.9646
& \textbf{0.8963} & \textbf{0.9623}
& 0.9228 & 0.8655
& 0.9219 & 0.995
& \textbf{0.993} & \textbf{1.000}
& \textbf{0.9963} & \textbf{1.000}
& \textbf{0.9395} & \textbf{0.983}
& 0.8792 & \textbf{1.000}
& \textbf{0.9030} & \textbf{0.933}
& 0.7359 & \textbf{0.9469} \\

\midrule
\toprule
\multirow{2}{*}{Method}
& \multicolumn{2}{c}{hinge2}
& \multicolumn{2}{c}{door2}
& \multicolumn{2}{c}{drawer2}
& \multicolumn{2}{c}{earphone\_case2}
& \multicolumn{2}{c}{fridge2}
& \multicolumn{2}{c}{injector2}
& \multicolumn{2}{c}{lamp2}
& \multicolumn{2}{c}{laptop2}
& \multicolumn{2}{c}{scissors2}
& \multicolumn{2}{c}{toilet\_seat2} \\
\cmidrule(lr){2-21}
& Pt & Obj & Pt & Obj & Pt & Obj & Pt & Obj & Pt & Obj
& Pt & Obj & Pt & Obj & Pt & Obj & Pt & Obj & Pt & Obj \\
\midrule

A0 (w/o $\theta$)
& 0.5720 & 0.5625
& 0.8723 & 0.6616
& 0.9393 & 0.418
& 0.6370 & 0.6666
& 0.8035 & 0.6194
& 0.6268 & 0.6265
& 0.6630 & 0.613
& 0.6450 & 0.560
& 0.6953 & 0.5283
& 0.5860 & 0.616 \\

A1 (Raw $\theta$)
& 0.5785 & 0.8947
& 0.9161 & 0.8333
& 0.9316 & 0.8400
& \textbf{0.8441} & 0.9170
& 0.9721 & 0.7790
& 0.6057 & 0.6083
& 0.8757 & 0.8867
& 0.6902 & 0.8183
& 0.8584 & 0.6767
& \textbf{0.8032} & 0.8567 \\

A2 (Fixed $\sin/\cos$)
& 0.5783 & 0.9699
& 0.9200 & 0.8533
& 0.9312 & 0.8567
& 0.8100 & 0.9650
& 0.9450 & \textbf{0.8774}
& 0.6075 & 0.5433
& \textbf{0.8783} & 0.9167
& 0.7010 & 0.8317
& 0.8584 & 0.7100
& 0.6978 & 0.8167 \\

A3 (Fourier $L{=}8$)
& 0.5850 & \textbf{0.9792}
& 0.8977 & 0.8467
& 0.9418 & \textbf{0.9183}
& 0.8259 & \textbf{0.9750}
& 0.9622 & 0.7774
& 0.6138 & \textbf{0.6350}
& 0.8416 & \textbf{0.9417}
& 0.6827 & 0.8733
& 0.9046 & \textbf{0.7350}
& 0.7403 & 0.8617 \\

A4 (Fourier $L{=}16$)
& \textbf{0.6862} & 0.9745
& 0.9173 & 0.7833
& 0.8998 & 0.8500
& 0.8038 & 0.9617
& 0.9617 & 0.7774
& 0.6156 & 0.4133
& 0.8377 & 0.9267
& \textbf{0.7731} & 0.8833
& 0.8906 & 0.6933
& 0.6855 & 0.8450 \\

SPA-SDF (full)
& 0.6012 & 0.9479
& \textbf{0.982} & \textbf{0.952}
& \textbf{0.9427} & 0.905
& 0.807 & 0.9717
& \textbf{0.9862} & 0.754
& \textbf{0.773} & 0.4701
& 0.8692 & 0.9267
& 0.7622 & \textbf{0.8933}
& \textbf{0.9065} & 0.733
& 0.762 & \textbf{0.883} \\

\midrule
\toprule
\multirow{2}{*}{Method}
& \multicolumn{2}{c}{hinge3}
& \multicolumn{2}{c}{door3}
& \multicolumn{2}{c}{drawer3}
& \multicolumn{2}{c}{earphone\_case3}
& \multicolumn{2}{c}{fridge3}
& \multicolumn{2}{c}{injector3}
& \multicolumn{2}{c}{lamp3}
& \multicolumn{2}{c}{laptop3}
& \multicolumn{2}{c}{scissors3}
& \multicolumn{2}{c}{toilet\_seat3} \\
\cmidrule(lr){2-21}
& Pt & Obj & Pt & Obj & Pt & Obj & Pt & Obj & Pt & Obj
& Pt & Obj & Pt & Obj & Pt & Obj & Pt & Obj & Pt & Obj \\
\midrule

A0 (w/o $\theta$)
& 0.5395 & 0.5717
& 0.9087 & 0.5383
& 0.9821 & 0.549
& 0.5417 & 0.516
& 0.8654 & 0.689
& \textbf{0.9637} & 0.604
& 0.7518 & 0.766
& 0.5789 & 0.585
& 0.6767 & 0.5283
& 0.7960 & 0.6233 \\

A1 (Raw $\theta$)
& 0.6776 & \textbf{0.9583}
& \textbf{0.9713} & 0.8667
& 0.9876 & 0.9267
& 0.6769 & 0.9617
& 0.9942 & 0.8267
& 0.8161 & 0.7250
& 0.9119 & 0.9650
& 0.6501 & \textbf{0.9317}
& 0.8497 & 0.6750
& 0.9271 & 0.9483 \\

A2 (Fixed $\sin/\cos$)
& 0.6980 & 0.9572
& 0.9633 & 0.7633
& 0.9888 & \textbf{0.9783}
& 0.6979 & 0.9783
& 0.9899 & 0.8500
& 0.8622 & \textbf{0.8967}
& \textbf{0.9261} & 0.9550
& 0.6501 & \textbf{0.9317}
& 0.8752 & 0.6833
& 0.9363 & 0.9100 \\

A3 (Fourier $L{=}8$)
& \textbf{0.6999} & \textbf{0.9583}
& 0.9555 & 0.7733
& 0.9894 & 0.9600
& \textbf{0.6989} & 0.9670
& 0.9897 & 0.8000
& 0.8311 & 0.8433
& 0.9231 & 0.9617
& 0.6783 & 0.9133
& \textbf{0.8929} & 0.6867
& \textbf{0.9418} & 0.9183 \\

A4 (Fourier $L{=}16$)
& 0.6635 & 0.9479
& 0.9619 & 0.7433
& \textbf{0.9929} & 0.9750
& 0.6760 & \textbf{0.9817}
& 0.9935 & 0.8500
& 0.8430 & 0.8617
& 0.9218 & \textbf{0.9667}
& 0.6900 & 0.9000
& 0.8725 & 0.6500
& 0.8845 & 0.9250 \\

SPA-SDF (full)
& 0.696 & 0.944
& 0.865 & \textbf{0.9992}
& 0.8935 & 0.6083
& 0.6678 & 0.917
& \textbf{0.9955} & \textbf{0.877}
& 0.9207 & 0.444
& 0.9018 & 0.956
& \textbf{0.6929} & 0.8833
& 0.863 & \textbf{0.765}
& 0.925 & \textbf{0.9567} \\

\midrule
\toprule
\multirow{2}{*}{Method}
& \multicolumn{2}{c}{usb}
& \multicolumn{2}{c}{usb2}
& \multicolumn{2}{c}{usb3}
& \multicolumn{2}{c}{utility\_knife}
& \multicolumn{2}{c}{utility\_knife2}
& \multicolumn{2}{c}{utility\_knife3}
& \multicolumn{2}{c}{folding\_knife}
& \multicolumn{2}{c}{folding\_knife2}
& \multicolumn{2}{c}{folding\_knife3}
& \multicolumn{2}{c}{Mean} \\
\cmidrule(lr){2-21}
& Pt & Obj & Pt & Obj & Pt & Obj & Pt & Obj & Pt & Obj
& Pt & Obj & Pt & Obj & Pt & Obj & Pt & Obj & Pt & Obj \\
\midrule

A0 (w/o $\theta$)
& 0.5110 & 0.5799
& 0.7290 & 0.5533
& 0.7962 & 0.483
& 0.8329 & 0.6989
& 0.7150 & 0.722
& 0.8265 & 0.4437
& 0.8097 & 0.626
& 0.7135 & 0.6166
& 0.8097 & 0.626
& 0.7196 & 0.6348 \\

A1 (Raw $\theta$)
& 0.9491 & 0.8104
& 0.8349 & 0.7483
& 0.7930 & 0.7717
& 0.8624 & 0.8817
& 0.7230 & 0.7915
& 0.8107 & 0.8859
& 0.9180 & 0.8806
& 0.8538 & 0.8683
& 0.8908 & 0.9267
& 0.8338 & 0.8500 \\

A2 (Fixed $\sin/\cos$)
& 0.9746 & 0.9750
& 0.8466 & \textbf{0.8317}
& 0.8710 & 0.8800
& 0.8202 & 0.8733
& 0.5948 & 0.8051
& 0.8239 & 0.8891
& 0.9229 & 0.8583
& 0.8321 & 0.8900
& 0.8818 & 0.9250
& 0.8356 & 0.8751 \\

A3 (Fourier $L{=}8$)
& 0.8489 & \textbf{0.9937}
& \textbf{0.8611} & 0.7883
& 0.8161 & \textbf{0.8883}
& 0.8356 & 0.8383
& 0.7295 & 0.7949
& 0.8335 & 0.8688
& 0.9208 & 0.8639
& 0.8652 & 0.8983
& 0.9133 & \textbf{0.9567}
& 0.8398 & 0.8781 \\

A4 (Fourier $L{=}16$)
& \textbf{0.9881} & 0.9854
& 0.8417 & 0.8200
& 0.8984 & 0.8800
& 0.8366 & 0.9167
& 0.6502 & 0.8695
& \textbf{0.8547} & \textbf{0.8922}
& \textbf{0.9410} & 0.8806
& 0.8664 & 0.8683
& 0.9032 & 0.9433
& 0.8431 & 0.8679 \\

SPA-SDF (full)
& 0.7923 & 0.9630
& 0.8276 & 0.7617
& \textbf{0.8988} & 0.7967
& \textbf{0.9728} & \textbf{0.9781}
& \textbf{0.7379} & \textbf{0.8831}
& 0.8390 & 0.8578
& 0.915 & \textbf{0.956}
& \textbf{0.8947} & \textbf{0.900}
& \textbf{0.915} & 0.956
& \textbf{0.8651} & \textbf{0.8843} \\

\bottomrule
\end{tabular}
\end{adjustbox}
\end{table*}

\begin{table*}[t]
\centering
\caption{Pose ablation on \textbf{unseen} classes. Best per column in bold.}

\label{tab:ablation2_unseen}
\footnotesize
\setlength{\tabcolsep}{2.2pt}
\begin{adjustbox}{width=\textwidth}
\begin{tabular}{l*{10}{cc}}
\toprule
\toprule
\multirow{2}{*}{Method}
& \multicolumn{2}{c}{hinge}
& \multicolumn{2}{c}{door}
& \multicolumn{2}{c}{drawer}
& \multicolumn{2}{c}{earphone\_case}
& \multicolumn{2}{c}{fridge}
& \multicolumn{2}{c}{injector}
& \multicolumn{2}{c}{lamp}
& \multicolumn{2}{c}{laptop}
& \multicolumn{2}{c}{scissors}
& \multicolumn{2}{c}{toilet\_seat} \\
\cmidrule(lr){2-21}
& Pt & Obj & Pt & Obj & Pt & Obj & Pt & Obj & Pt & Obj
& Pt & Obj & Pt & Obj & Pt & Obj & Pt & Obj & Pt & Obj \\
\midrule

A0 (w/o $\theta$)
& 0.5683 & 0.595
& 0.6676 & 0.6703
& \textbf{0.9442} & 0.441
& 0.8088 & 0.901
& 0.8786 & 0.694
& 0.9535 & 0.971
& 0.7120 & 0.813
& 0.6903 & 0.5967
& 0.6746 & 0.693
& 0.5480 & 0.672 \\

A1 (Raw $\theta$)
& 0.7711 & 0.8925
& \textbf{0.8624} & 0.8852
& 0.9156 & 0.7826
& 0.8836 & 0.9017
& \textbf{0.9911} & 0.9919
& 0.6260 & 0.4857
& 0.8937 & 0.9183
& 0.8160 & 0.9097
& \textbf{0.9029} & 0.9433
& 0.6785 & 0.8912 \\

A2 (Fixed $\sin/\cos$)
& 0.9237 & 0.9338
& \textbf{0.8624} & 0.9000
& 0.9335 & 0.7633
& \textbf{0.9009} & 0.9250
& 0.9905 & \textbf{1.0000}
& 0.6131 & 0.4536
& 0.8761 & 0.9583
& 0.8306 & 0.9371
& 0.8961 & 0.9100
& \textbf{0.7151} & 0.9088 \\

A3 (Fourier $L{=}8$)
& 0.8270 & \textbf{0.9500}
& 0.6934 & 0.7704
& 0.8983 & 0.6430
& 0.8951 & \textbf{0.9780}
& 0.9850 & 0.9940
& \textbf{0.9680} & \textbf{0.9847}
& \textbf{0.9050} & 0.9170
& 0.7435 & 0.9177
& 0.8810 & 0.7783
& 0.6629 & 0.7471 \\

A4 (Fourier $L{=}16$)
& 0.9239 & 0.9437
& 0.7919 & 0.7296
& 0.9028 & 0.6383
& 0.8753 & 0.9333
& 0.9861 & 0.9774
& 0.6048 & 0.3982
& 0.8802 & 0.9333
& \textbf{0.8394} & \textbf{0.9694}
& 0.8583 & 0.9433
& 0.6737 & 0.8632 \\

SPA-SDF (full)
& \textbf{0.9599} & 0.9388
& 0.8489 & \textbf{0.9444}
& 0.9396 & \textbf{0.8600}
& 0.8749 & 0.9317
& \textbf{0.9911} & 0.9887
& 0.6000 & 0.6304
& 0.8959 & \textbf{0.9800}
& 0.7535 & 0.9226
& 0.8844 & \textbf{0.9600}
& 0.7091 & \textbf{0.9132} \\

\midrule
\toprule
\multirow{2}{*}{Method}
& \multicolumn{2}{c}{hinge2}
& \multicolumn{2}{c}{door2}
& \multicolumn{2}{c}{drawer2}
& \multicolumn{2}{c}{earphone\_case2}
& \multicolumn{2}{c}{fridge2}
& \multicolumn{2}{c}{injector2}
& \multicolumn{2}{c}{lamp2}
& \multicolumn{2}{c}{laptop2}
& \multicolumn{2}{c}{scissors2}
& \multicolumn{2}{c}{toilet\_seat2} \\
\cmidrule(lr){2-21}
& Pt & Obj & Pt & Obj & Pt & Obj & Pt & Obj & Pt & Obj
& Pt & Obj & Pt & Obj & Pt & Obj & Pt & Obj & Pt & Obj \\
\midrule

A0 (w/o $\theta$)
& 0.5690 & 0.5648
& 0.8660 & 0.4390
& 0.9276 & 0.4000
& 0.5183 & 0.6574
& 0.8320 & 0.6131
& 0.7556 & \textbf{0.6192}
& 0.6120 & 0.5566
& 0.6636 & 0.5360
& 0.7149 & 0.4466
& 0.4370 & 0.6433 \\

A1 (Raw $\theta$)
& 0.5571 & 0.8610
& \textbf{0.9328} & 0.8433
& 0.9266 & \textbf{0.9300}
& 0.8303 & 0.9733
& \textbf{0.9795} & 0.8562
& 0.6246 & 0.4350
& \textbf{0.8851} & 0.8733
& 0.7112 & 0.9083
& 0.8714 & 0.6867
& 0.7497 & \textbf{0.8633} \\

A2 (Fixed $\sin/\cos$)
& 0.5665 & 0.9618
& 0.9278 & 0.7867
& 0.9335 & 0.7633
& 0.8328 & 0.9667
& 0.9592 & \textbf{0.8688}
& 0.6281 & 0.4183
& 0.8699 & 0.8900
& 0.7121 & 0.9083
& 0.8920 & 0.7433
& 0.6731 & 0.7717 \\

A3 (Fourier $L{=}8$)
& 0.5339 & 0.9120
& 0.7348 & \textbf{0.9745}
& 0.9370 & 0.7317
& 0.7510 & 0.9683
& 0.9680 & 0.7500
& \textbf{0.7879} & 0.5433
& 0.7394 & 0.8950
& 0.7210 & 0.8650
& 0.8744 & 0.7000
& 0.6658 & 0.7200 \\

A4 (Fourier $L{=}16$)
& 0.5485 & \textbf{0.9711}
& 0.9270 & 0.6833
& 0.9028 & 0.6383
& 0.8354 & 0.9517
& 0.9679 & 0.7812
& 0.6298 & 0.3533
& 0.8659 & 0.8667
& \textbf{0.7547} & 0.8683
& 0.8754 & 0.7233
& 0.6795 & 0.8133 \\

SPA-SDF (full)
& \textbf{0.5949} & \textbf{0.9711}
& 0.9159 & 0.7533
& \textbf{0.9396} & 0.8600
& \textbf{0.8417} & \textbf{0.9833}
& 0.9707 & 0.8391
& 0.6287 & 0.5583
& 0.8623 & \textbf{0.9050}
& 0.6903 & \textbf{0.9167}
& \textbf{0.9090} & \textbf{0.7800}
& \textbf{0.7900} & 0.8550 \\

\midrule
\toprule
\multirow{2}{*}{Method}
& \multicolumn{2}{c}{hinge3}
& \multicolumn{2}{c}{door3}
& \multicolumn{2}{c}{drawer3}
& \multicolumn{2}{c}{earphone\_case3}
& \multicolumn{2}{c}{fridge3}
& \multicolumn{2}{c}{injector3}
& \multicolumn{2}{c}{lamp3}
& \multicolumn{2}{c}{laptop3}
& \multicolumn{2}{c}{scissors3}
& \multicolumn{2}{c}{toilet\_seat3} \\
\cmidrule(lr){2-21}
& Pt & Obj & Pt & Obj & Pt & Obj & Pt & Obj & Pt & Obj
& Pt & Obj & Pt & Obj & Pt & Obj & Pt & Obj & Pt & Obj \\
\midrule

A0 (w/o $\theta$)
& 0.5002 & 0.5600
& 0.8520 & 0.5830
& 0.9680 & 0.5416
& 0.5380 & 0.6366
& 0.8965 & 0.5439
& 0.9652 & 0.5829
& 0.7647 & 0.6560
& 0.5590 & 0.5083
& 0.6635 & 0.4400
& 0.7983 & 0.5533 \\

A1 (Raw $\theta$)
& 0.6036 & \textbf{0.9572}
& \textbf{0.9250} & 0.7233
& 0.9843 & 0.9467
& 0.7658 & 0.9483
& \textbf{0.9902} & \textbf{0.8817}
& 0.7538 & 0.7567
& 0.9127 & \textbf{0.9667}
& 0.6730 & 0.8683
& 0.8721 & 0.7750
& 0.9115 & \textbf{0.9433} \\

A2 (Fixed $\sin/\cos$)
& \textbf{0.6398} & 0.9387
& 0.8955 & 0.7817
& \textbf{0.9883} & \textbf{0.9850}
& 0.7926 & 0.9433
& 0.9899 & 0.8800
& 0.8166 & \textbf{0.9083}
& 0.9339 & 0.9617
& \textbf{0.7063} & \textbf{0.9483}
& 0.8849 & 0.7500
& 0.9049 & 0.8983 \\

A3 (Fourier $L{=}8$)
& 0.6201 & 0.8796
& 0.8159 & \textbf{0.9811}
& 0.9580 & 0.5600
& 0.7041 & 0.8833
& 0.9770 & 0.7190
& \textbf{0.9890} & 0.4770
& 0.8754 & 0.8750
& 0.5573 & 0.7433
& 0.8741 & 0.7510
& 0.8417 & 0.9022 \\

A4 (Fourier $L{=}16$)
& 0.6052 & 0.9468
& 0.9067 & 0.7167
& 0.9828 & 0.9267
& \textbf{0.7931} & 0.9233
& 0.9873 & 0.8500
& 0.7999 & 0.7717
& 0.9213 & 0.9400
& 0.6929 & 0.8083
& 0.8577 & 0.7933
& 0.8842 & 0.9117 \\

SPA-SDF (full)
& 0.6255 & 0.9340
& 0.9123 & 0.7783
& 0.9851 & \textbf{0.9850}
& 0.7764 & \textbf{0.9650}
& 0.9891 & 0.8533
& 0.7767 & 0.8450
& \textbf{0.9382} & 0.9617
& 0.6933 & 0.8850
& \textbf{0.8912} & \textbf{0.8333}
& \textbf{0.9364} & 0.9133 \\

\midrule
\toprule
\multirow{2}{*}{Method}
& \multicolumn{2}{c}{usb}
& \multicolumn{2}{c}{usb2}
& \multicolumn{2}{c}{usb3}
& \multicolumn{2}{c}{utility\_knife}
& \multicolumn{2}{c}{utility\_knife2}
& \multicolumn{2}{c}{utility\_knife3}
& \multicolumn{2}{c}{folding\_knife}
& \multicolumn{2}{c}{folding\_knife2}
& \multicolumn{2}{c}{folding\_knife3}
& \multicolumn{2}{c}{Mean} \\
\cmidrule(lr){2-21}
& Pt & Obj & Pt & Obj & Pt & Obj & Pt & Obj & Pt & Obj
& Pt & Obj & Pt & Obj & Pt & Obj & Pt & Obj & Pt & Obj \\
\midrule

A0 (w/o $\theta$)
& 0.8130 & 0.5960
& 0.6000 & 0.6947
& 0.7810 & 0.5500
& \textbf{0.8196} & 0.8693
& 0.6973 & 0.6983
& 0.8362 & 0.5031
& 0.8218 & 0.5630
& 0.7201 & 0.6083
& 0.8218 & 0.5630
& 0.7373 & 0.6095 \\

A1 (Raw $\theta$)
& 0.9475 & 0.8902
& 0.8058 & 0.7367
& 0.7836 & 0.6433
& 0.8028 & 0.8790
& 0.7200 & \textbf{0.7850}
& 0.8203 & 0.8484
& 0.9204 & 0.7527
& 0.8662 & 0.9167
& 0.9095 & 0.9067
& 0.8301 & 0.8450 \\

A2 (Fixed $\sin/\cos$)
& 0.9795 & 0.9646
& 0.8292 & 0.6950
& 0.8548 & 0.8533
& 0.7511 & 0.8694
& 0.5774 & 0.7500
& 0.8322 & 0.8391
& \textbf{0.9242} & 0.8649
& 0.8238 & 0.9017
& 0.9050 & 0.8950
& 0.8350 & 0.8563 \\

A3 (Fourier $L{=}8$)
& 0.9540 & 0.9630
& 0.8012 & 0.6550
& 0.8576 & 0.6550
& 0.7761 & \textbf{0.9750}
& \textbf{0.7231} & 0.7217
& 0.8426 & 0.7766
& 0.9161 & \textbf{0.9130}
& 0.8579 & 0.8000
& 0.9161 & \textbf{0.9130}
& 0.8212 & 0.8175 \\

A4 (Fourier $L{=}16$)
& \textbf{0.9891} & 0.9585
& 0.8200 & 0.7067
& \textbf{0.8719} & 0.7917
& 0.8127 & 0.8823
& 0.6337 & 0.6517
& \textbf{0.8601} & 0.8281
& 0.9241 & 0.8189
& \textbf{0.8785} & 0.8983
& 0.9184 & 0.9017
& 0.8324 & 0.8207 \\

SPA-SDF (full)
& 0.9859 & \textbf{0.9829}
& \textbf{0.8364} & \textbf{0.8017}
& 0.8061 & \textbf{0.8817}
& 0.7680 & 0.7823
& 0.7165 & 0.6017
& 0.8373 & \textbf{0.9297}
& 0.9179 & 0.8541
& 0.8741 & \textbf{0.9267}
& \textbf{0.9274} & 0.8900
& \textbf{0.8410} & \textbf{0.8740} \\

\bottomrule
\end{tabular}
\end{adjustbox}
\end{table*}

\end{document}